\documentclass[conference,compsoc]{IEEEtran}

\usepackage{pifont}

\ifCLASSOPTIONcompsoc
  \usepackage[nocompress]{cite}
\else
  \usepackage{cite}
\fi

\ifCLASSINFOpdf
  \usepackage[pdftex]{graphicx}
\else
\fi

\usepackage{amsmath}
\usepackage{amssymb}

\usepackage{makecell}
\usepackage{multirow}
\usepackage{booktabs}

\usepackage{url}
\usepackage[colorlinks=true,
    linkcolor=blue,
    citecolor=blue,
    urlcolor=blue,
    filecolor=blue]{hyperref}

\usepackage{xcolor}

\newif\ifdraft
\draftfalse

\newcommand{\warnreader}[1]{\noindent\textcolor{red}{\textsc{Warning: #1}}}

\newcommand{\parhead}[1]{\medskip\noindent\textbf{#1}}

\newcommand{\figref}[1]{Fig.~\ref{#1}}

\newcommand{\figrefs}[2]{Figs.~\ref{#1}--\ref{#2}}
\newcommand{\tabref}[1]{Tab.~\ref{#1}}
\newcommand{\secref}[1]{\S\ref{#1}}
\newcommand{\secrefs}[2]{\S\ref{#1}--\ref{#2}}
\newcommand{\appref}[1]{App.~\ref{#1}}

\newcommand{\eqnref}[1]{Eq.~\ref{#1}}
\renewcommand{\eqref}[1]{Eq.~\ref{#1}}
\newcommand{\secreftable}[1]{\S\ref{#1}} %

\newcommand{\unsafeatk}[0]{Unsafe@k}
\newcommand{\cmark}{\ding{51}} %
\newcommand{\xmark}{\ding{55}} %
\newcommand{\safegen}[0]{NSFW-1k}
\newcommand{\ftbased}[0]{fine-tuning-based}
\newcommand{\ours}[0]{\textbf{D}istillation-\textbf{I}ntegrated \textbf{C}oncept \textbf{E}rasure (\textbf{DICE})}
\newcommand{\fluxsch}[0]{FLUX Schnell}
\newcommand{\fluxdev}[0]{FLUX Dev}
\newcommand{\fluxscfm}[0]{FLUX SCFM}
\newcommand{\sdlarge}[0]{SD3.5 Large}
\newcommand{\sdturbo}[0]{SD3.5 Turbo}

\begin{document}
\title{Haste Makes Waste? Towards Safe Few-Step Text-to-Image Models}

\author{
\IEEEauthorblockN{
  Lior Ziv %
  and
  Mahmood Sharif %
}
\IEEEauthorblockA{
\textit{Tel Aviv University}\\
Email:
\href{mailto:lz1@mail.tau.ac.il}{lz1@mail.tau.ac.il},
\href{mailto:mahmoods@tauex.tau.ac.il}{mahmoods@tauex.tau.ac.il}
}
}

\maketitle

\begin{abstract}

While few-step distilled text-to-image models have gained traction due to their fast generation times, the speed-up implications on their unsafe generation rates remain unknown. We uncover an inherent tension between speed and safety: few-step models are often more unsafe than their teachers, and reducing the number of inference steps consistently raises unsafe generation rates. We further find that acceleration makes unsafe concepts harder to erase in few-step models, with existing concept-erasure methods markedly less effective than in teachers.
We offer a mechanistic diagnosis for why erasure methods fail on few-step models, finding that the fragility of few-step models' generation process and the way generation guidance is usually distilled into those models may hinder established concept-erasure methods. 
To address the speed-safety trade-off, we propose \textbf{DICE} (\emph{Distillation-Integrated Concept Erasure}), a post-distillation erasure method tailored for few-step models. DICE aligns both its latent construction and its generation-preservation loss with the distillation method that produced the model, keeping erasure faithful to few-step models. DICE demonstrates strong results in our evaluation. For example, on FLUX Schnell, DICE reduces unsafe-generation rate to 0.046 (vs.\ $\ge$0.161 for prior erasure method we test), while mostly preserving benign generation quality.\footnote{Our code and models are available at: \url{https://github.com/plus-tau/safe-distilled-t2i}.}

\end{abstract}

\IEEEpeerreviewmaketitle

\medskip{}
\warnreader{This paper contains explicit text and (censored) images that readers may be uncomfortable with.}

\section{Introduction}
\label{sec:introduction}

Text-to-image (T2I) diffusion models~\cite{rombach2022high,stabilityai2024sd35,xie2024sana,flux2024} are increasingly deployed in user-facing products, making the safety of the content they generate a first-order deployment concern.
A dominant deployment trend over the past two years has been a move from many-step samplers to fast, distilled few-step variants---e.g., \sdturbo{}~\cite{stabilityai2024sd35,sauer2024adversarial}, Sana Sprint~\cite{chen2025sana,sauer2024fast}, \fluxsch{}~\cite{flux2024}, and SCFM-style distilled models~\cite{cai2026shortcutting}---which generate images in one to four steps and are routinely shipped as drop-in replacements for their slower teachers.
However, whether this acceleration preserves the safety properties of the underlying model has not been systematically audited. Prior work on T2I safety benchmarking~\cite{schramowski2023safe,li2024safegen,kim2023towards} and on concept erasure (e.g., ESD~\cite{gandikota2023erasing}, UCE~\cite{gandikota2024unified}, EraseAnything~\cite{gao2025eraseanything}, Meta-Unlearning~\cite{gao2025meta}) was developed and evaluated against full-step teacher models, and two questions of immediate deployment relevance remain open: \emph{(i)}~how does the safety profile of a T2I system change when it is shipped through the acceleration pipeline? and \emph{(ii)}~do existing concept-erasure defenses, applied at the natural insertion points of that pipeline, continue to function once the deployment target is a few-step distilled model?
This paper sets to answer these questions.

We find that acceleration is not safety-neutral---few-step acceleration and reducing T2I models' inference step count both result in higher rates of unsafe content generation (\secrefs{subsec:safety_under_acceleration}{subsec:step_count_is_a_hidden_safety_variable}).
We then turn to checking whether existing concept-erasure methods originally designed for full-step baseline models are effective on few-step distilled models (\secref{subsec:erasure_integration_under_acceleration}). The results show that applying these to few-step models either directly (post-distillation), prior to distillation, or via transfer from the teacher is significantly less effective than applying them to full-step models.

To better understand why erasure fails in few-step models, we carry a mechanistic study (\secref{sec:mechanistic_diagnosis}).
A simple explanation would be that acceleration and concept erasure directly compete.
However, we find that the update directions of acceleration and erasure techniques are mostly orthogonal, suggesting that other issues hinder erasure within few-step models.
Complementing prior work on the geometry and early decision concentration of few-step models \cite{huangrestoring,gandikota2026distilling}, our empirical analyses uncover two failure modes that emerge when full-step teachers are distilled into one-to-four-step students.
First, few-step students are more brittle under small changes: relative to their teachers, perturbing either the latent input or the model weights slightly can drastically change the student's next generation update. Second, the way guidance is baked into the weights of few-step models distorts the concept direction used during erasure: the student direction is rotated away from the teacher's and inflated in norm, driving erasure updates to a degraded target. Together, these findings suggest that existing erasure methods fail not because safety and speed are inherently incompatible, but because post-hoc edits are applied to models whose few-step generation behavior differs from the full-step models those methods were designed for.

In an attempt to address the gap,
we present \textbf{DICE} (\emph{Distillation-Integrated Concept Erasure})---to our knowledge, the first concept-erasure method tailored for few-step models.
DICE retargets two key components of the standard erasure so they are better fit for the distilled student: \emph{(i)}~the generation-preservation loss keeps the edited model's output close to the original distilled model's, rather than the standard flow-matching training target that acceleration methods may deviate from; and \emph{(ii)}~the noisy latent on which the ESD loss is evaluated is sampled along the straight noise-to-image line between Gaussian noise and a clean concept image generated by the frozen student---mirroring the latent construction used in distillation rather than the deployment scheduler grid where erasure direction may be degraded and where the few-step model is sensitive to minor perturbations.
We instantiate DICE on \fluxsch{}, \sdturbo{} and \fluxscfm{}, evaluate it on safety, speed, and quality axes, and run ablations that isolate the two design choices.
Across the methods we evaluate, DICE is the only configuration that simultaneously significantly reduces unsafe concept generation (pixel-level detectors and \unsafeatk{}, with qualitative confirmation) and maintains general image quality on already-distilled rectified-flow models.

We summarize our core contributions as follows:
\begin{itemize}
    \item We show that T2I acceleration is not safety-neutral: a trade-off exists between safety and speed in current open-weights flow-matching models (\secrefs{subsec:safety_under_acceleration}{subsec:step_count_is_a_hidden_safety_variable}).
    \item We show that current concept-erasure methods do not effectively erase unsafe concepts from few-step models (\secref{subsec:erasure_integration_under_acceleration}); our mechanistic study helps shed light on why these methods may fail (\secref{sec:mechanistic_diagnosis}).
    \item Motivated by our findings, we present DICE, an erasure method tailored for few-step models that is able to better suppress unsafe concepts while preserving image quality in these models (\secref{sec:dice}).
\end{itemize}

\section{Background}
\label{sec:background}

\subsection{Flow-Matching T2I Models}

Recent T2I models increasingly adopt flow-matching or rectified-flow formulations. These models operate in the latent space of a pretrained Variational Autoencoder (VAE). They learn a velocity field that defines a generative path from a noise sample $x_1$ to a clean image latent $x_0$, and generation is performed by numerically integrating this learned velocity field. A solver timestep $t$ indexes a noise level $\sigma_t$ from the model's noise schedule, with $\sigma_t \approx 1$ near the noisy endpoint and $\sigma_t \approx 0$ near the clean endpoint. For a text condition $c$ and model parameters $\theta$, we write the predicted velocity as $v_\theta(z_t,c,t)$. The corresponding noisy latent can be written as the linear interpolation
\begin{equation}
\label{eq:flow_interpolation}
z_t
=
(1-\sigma_t)x_0 + \sigma_t x_1
=
x_0 + \sigma_t(x_1-x_0)
\end{equation}
where $x_1$ is Gaussian noise.

The learned velocity field specifies how to move along this path during sampling. Starting from noise $z_{t=1}=x_1$, the sampler repeatedly evaluates $v_\theta(z_t,c,t)$ and takes numerical integration steps toward lower noise levels, producing an estimate of the clean latent at $t=0$. During flow-matching training, the model is supervised to predict the reverse denoising velocity of the interpolation in \eqref{eq:flow_interpolation}: since $\partial z_t / \partial \sigma_t = x_1-x_0$, the target direction for decreasing noise is $x_1-x_0$. Thus the denoiser learns the direction that transports noisy latents back toward clean image latents, which are then decoded to images using the VAE decoder.

When the endpoints and timestep are clear from context, we write a noisy latent simply as $z$, omitting the particular $x_0$, $x_1$, and $t$ used to construct it. We use this notation to distinguish two ways of obtaining latents $z$ for analysis or erasure training. When $z$ is synthetic and obtained by running a particular model for one or more solver steps along its own sampling trajectory, we call it \emph{on-policy} for that model. When $z$ is constructed outside that rollout, by taking a clean image $x_0$ and re-noising it to a chosen $\sigma_t$ using \eqref{eq:flow_interpolation}, we call it \emph{off-policy}.

Compared to classical DDPM/DDIM-style \cite{song2020denoising} diffusion sampling, rectified-flow formulations are designed to support straighter trajectories and fewer integration steps. However, these models still commonly require many solver steps at inference. Two prominent open examples are \fluxdev{} and \sdlarge{} \cite{bfl_flux_announcement, stabilityai2024sd35}. \sdlarge{} uses standard classifier-free guidance (CFG) \cite{ho2022classifier}: the model is evaluated with both the text condition $c$ and the null condition $\varnothing$, and the guided velocity is formed as
\[
v_{\mathrm{CFG}}(z,c,t;\omega)
=
v_\theta(z,\varnothing,t)
+
\omega\left(v_\theta(z,c,t)-v_\theta(z,\varnothing,t)\right),
\]
where $\omega$ is the guidance scale. This keeps the guidance operation external to the model weights and preserves separate conditional and unconditional predictions at inference.

\fluxdev{} handles guidance differently: it conditions the model on a guidance scalar and approximates guided generation in a single forward pass:
\[
v_{\mathrm{CFG}}(z,c,t;\omega)
=
v_\theta(z,c,t,\omega)
\]
We refer to this as guidance distillation, and distinguish it from few-step distillation. Guidance distillation changes how CFG is represented, while few-step distillation changes the number and size of solver steps the model is trained to support. \fluxdev{} is guidance-distilled from the closed-weights FLUX Pro, but it is not a few-step model, and is sampled with many inference steps, 50 by default.

\subsection{Few-Step Models}
\label{subsec:few-step-models}

To accelerate generation, few-step distillation trains a student model to generate high-quality images in only 1--4 inference steps. In many modern T2I settings, the student shares the teacher architecture and is initialized from the teacher, but its training objective differs from the original flow-matching objective: instead of learning only the local velocity target used to train the teacher, the student is optimized to reproduce the teacher's behavior under much larger sampling steps.

\parhead{Distillation methods} We study several forms of few-step distillation. Adversarial Diffusion Distillation (ADD) combines teacher-based distillation with an adversarial objective, encouraging the student to produce realistic samples in the aggressive 1--4 step regime. Latent Adversarial Diffusion Distillation (LADD) performs adversarial distillation in the latent space of a pretrained latent diffusion model, using synthetic teacher-generated data. In T2I applications, ADD/LADD students are trained to match a guided teacher and are then sampled without external CFG. We refer to this as guidance ``bake-in,'' as the CFG behavior is baked into the student's weights. As a result, a particular teacher-guidance regime is effectively absorbed into the student. Shortcut Consistency Flow Matching (SCFM), by contrast, is a consistency-based method for flow-matching models. It builds shortcut targets by composing two smaller velocity steps over a longer interval. The objective contains both a teacher-guided term, where the two-step target is computed using the frozen pretrained flow model, and a self-distillation term, where the target is computed using an Exponential Moving Average (EMA) or online student. Thus, the frozen teacher remains a fixed anchor throughout training, while the EMA term encourages internal shortcut consistency. In this sense, SCFM is a lighter, progressive-like form of acceleration: the original SCFM work targets moderate few-step sampling in the 3--8 step regime with substantially less training time than ADD/LADD-style distillation, while staying closer to the teacher's velocity-space objective.

This variety is important for our work. \sdturbo{} is described as an adversarially distilled variant of \sdlarge{}, trained using ADD \cite{stabilityai2024sd35}. \fluxsch{} is publicly described as a few-step model trained using LADD, but the specific teacher model and full distillation recipe are not disclosed \cite{bfl_flux_schnell_modelcard, bfl_flux_announcement}. Thus, unlike SD3.5 Turbo, \fluxsch{} does not provide a known teacher-student pair for controlled analysis. We therefore apply SCFM to \fluxdev{} to obtain \fluxscfm{}, a controlled FLUX-based teacher-student pair with a known teacher. Overall, we evaluate released few-step checkpoints together with \fluxscfm{}, covering adversarial, latent-adversarial, and consistency/shortcut distillation.

\subsection{Concept Erasure}

Concept erasure aims to remove an undesirable concept from a pretrained T2I model while preserving its ability to generate unrelated content. Erased concepts may correspond to unsafe content, copyrighted styles, memorized identities, or other concepts that a model provider wants to suppress.

A prominent family of erasure methods fine-tunes the model by optimizing gradient-based objectives. A central building block in this family is the ESD loss, introduced by Erased Stable Diffusion (ESD) and reused by many later fine-tuning-based erasure methods \cite{gandikota2023erasing,gao2025meta,gao2025eraseanything,kim2023towards,zhang2024defensive,huang2024receler}. $\mathcal{L}_{\mathrm{ESD}}$ suppresses a target concept by modifying the model's conditional prediction relative to its unconditional prediction. In velocity-prediction notation, it can be written as
\begin{equation}
\begin{aligned}
\label{eq:esd}
\mathcal{L}_{\mathrm{ESD}}
={}&
\Bigl\|
v_\theta(z,c,t)
-
\Bigl[
v_{\theta^\star}(z,\varnothing,t) \\
&\quad
-\eta
\bigl(
v_{\theta^\star}(z,c,t)
-
v_{\theta^\star}(z,\varnothing,t)
\bigr)
\Bigr]
\Bigr\|_2^2 
\end{aligned}
\end{equation}
where $v_{\theta^\star}$ is the frozen pretrained model, $v_\theta$ is the edited model, $c$ is the erased concept, $\varnothing$ is the null condition, and the negative guidance strength $\eta$ controls the strength of the negative signal. Intuitively, the method pushes the edited model away from the concept direction encoded by the difference between conditional and unconditional predictions.

Recent %
methods for fine-tuning T2I models have adopted Low-Rank Adaptation (LoRA)~\cite{hu2022lora}. Instead of updating the full weight matrix ($\theta_\ell$) of each layer ($\ell$), LoRA keeps the pretrained weights frozen and learns a low-rank residual update,
\begin{equation}
\label{eq:lora}
\theta_\ell \leftarrow \theta_\ell + \Delta\theta_\ell,
\qquad
\Delta\theta_\ell = B_\ell A_\ell ,
\end{equation}
where $A_\ell \in \mathbb{R}^{r \times d_{\mathrm{in}}}$ and $B_\ell \in \mathbb{R}^{d_{\mathrm{out}} \times r}$ are trainable low-rank matrices, with $r \ll \min(d_{\mathrm{in}}, d_{\mathrm{out}})$. This parameterization has become popular due to its simplicity and effectiveness, as well as its ability to substantially reduce the number of optimized parameters, enabling faster and more memory-efficient training. Since the original model weights remain fixed and only the residual update ($\Delta\theta_\ell$) is optimized, LoRA can also help preserve the pretrained model and mitigate catastrophic forgetting during fine-tuning.

Not all erasure methods are \ftbased{}. We also include UCE \cite{gandikota2024unified} and SPEED \cite{li2026speed} as closed-form editing methods. UCE analytically edits the model's text-conditioning projections in order to modify how specific text concepts are mapped into the image-generation process, without iterative fine-tuning. SPEED refines UCE to enable multiple concept erasure and precise edits specific to target concepts. In older U-Net-based \cite{ronneberger2015u} diffusion models these edits target cross-attention projections; in transformer-based models such as FLUX and SD3.5, the corresponding edits apply to the initial text-conditioning projection layers rather than to cross-attention modules.

Most concept-erasure methods were designed and evaluated primarily on SD 1.x-style latent diffusion models with U-Net backbones, DDPM/DDIM-style denoising objectives, and standard CFG sampling. Their assumptions do not necessarily transfer cleanly to modern T2I models such as FLUX and SD3.5, which use Multimodal Diffusion Transformer (MMDiT) \cite{esser2024scaling} backbones and flow-matching velocity prediction rather than DDPM/DDIM-style noise prediction. A recent work, EraseAnything, explicitly targets concept erasure in newer flow-matching T2I models and shows that adapting erasure to these architectures requires additional care \cite{gao2025eraseanything}. We therefore evaluate both classical erasure methods and EraseAnything.

\subsection{Threat Model}
We consider a defender that modifies a pretrained, potentially distilled T2I model to prevent the generation of prohibited visual content while preserving its benign generation quality and inference speed. In line with prior work (e.g.,~\cite{gandikota2023erasing,gandikota2024unified,gao2025eraseanything}), we primarily assume a black-box adversary who controls input prompts and seeks to elicit prohibited content. In evaluations of adaptive adversaries (\secref{subsec:adaptive_adv_eval}), we further consider adversaries with white-box access to the text encoder and to the full model weights.

\section{Empirical Safety Trade-offs}
\label{sec:findings}

We begin by exploring the safety implications of few-step acceleration, contrasting few-step students with full-step teacher models (\secref{subsec:safety_under_acceleration}), and assessing how reducing the number of generation steps impacts safety (\secref{subsec:step_count_is_a_hidden_safety_variable}). 
Our results show a consistent trade-off between speed and safety: faster generation (i.e., lower number of generation steps) is associated with more unsafe content emitted by the models. We also apply popular and leading erasure methods~\cite{gandikota2023erasing,gao2025eraseanything,gandikota2024unified} to few-step models and check their impact on safety, speed, and utility (\secref{subsec:erasure_integration_under_acceleration}). We find that erasure methods demonstrate significantly lower efficacy over all the few-step models we test. We further test alternative strategies to integrate few-step acceleration and erasures, such as erasing a teacher before distillation (\secref{subsec:erasure_integration_under_acceleration}). Across all strategies, erasure remains consistently weaker in the resulting few-step models than in their full-step counterparts, while introducing additional strategy-specific weaknesses. Before describing our findings, we present the experimental setup (\secref{subsec:experimental_setup}).

\subsection{Experimental Setup}
\label{subsec:experimental_setup}

\parhead{Models}
For our experiments, we focus on rectified flow-matching models, which have emerged as a dominant formulation in recent state-of-the-art T2I systems \cite{esser2024scaling}, specifically the popular FLUX and SD3.5 model families. Both have few-step variants produced using ADD/LADD \cite{sauer2024adversarial, sauer2024fast}---\sdturbo{} which is based on \sdlarge{} \cite{stabilityai2024sd35}, and \fluxsch{} which is produced using LADD \cite{sauer2024fast,flux2024}, but the specific teacher model used is not disclosed. Thus, we employ the SCFM distillation method \cite{cai2026shortcutting} over \fluxdev{} to produce \fluxscfm{}. We use the SCFM training recipe and parameters recommended by Cai et al.\ \cite{cai2026shortcutting}---full details are in \appref{subsec:appendix_flux_scfm_training_details}.
By training \fluxscfm{}, we effectively create a known teacher-student pair for the FLUX model family.
Because SCFM is a distillation method based on the self-consistency loss \cite{frans2025one} rather than an adversarial-based loss like ADD or LADD, \fluxscfm{} helps us corroborate our findings on a profoundly different distillation approach.
Since few-step distillation is a resource-intensive process compared to concept erasure or inference, and since we already have a known open-weights student-teacher pair in the SD3.5 family, we do not apply SCFM training for the SD3.5 family. Another approach for producing a student for FLUX would be applying LADD ourselves, but the LADD process is significantly more resource intensive \cite{cai2026shortcutting} than SCFM, making it infeasible to run. 

We use each model through its official implementation in the \texttt{diffusers} library, together with default components it supplies: the Euler Discrete scheduler and its timestep schedule, the image variational autoencoder (VAE), and the corresponding text tokenizer and encoder. 
Unless stated otherwise, we use the recommended default inference configuration for each model \cite{flux2024,cai2026shortcutting,stabilityai2024sd35}: CFG scale 3.5 for SD3.5 and FLUX, CFG scale 5 for \fluxscfm{}, which lies within its recommended operating range [4.5, 6], 28 inference steps for \sdlarge{}, 50 inference steps for \fluxdev{}, and 4 inference steps for all few-step models. 

\parhead{Unsafe prompts and safety checkers}
Following common practice \cite{chen2025comprehensive,schramowski2023safe,gandikota2023erasing}, we assess nudity and violence erasure, with nudity being our main focus. Due to a lack of explicitness we encounter in the widely used I2P \cite{schramowski2023safe} dataset, we follow multiple recent papers \cite{yang2025certphash,jiang2026jailbreaking} and use the NSFW-56k \cite{li2024safegen} dataset for nudity, which we find significantly more explicit. We report results on a randomly selected subset of 1k prompts from the NSFW-56k dataset (hereafter, \safegen{}). We also report results on the I2P dataset, adopted due to its popularity in the field \cite{gandikota2023erasing,gao2025eraseanything,huang2024receler} in \appref{subsec:appendix_dice_results_on_i2p}. 
To evaluate the safety of generated images in the nudity use-case, we leverage two widely used detectors: NudeNet \cite{nudenet}, a CNN-based detector, and SD-Checker \cite{rando2022red}, a CLIP-based detector shipped with earlier Stable Diffusion releases. We find the two checkers to be mutually consistent, and both have been shown to agree with human annotators \cite{qu2025unsafebench}. 
\appref{subsec:appendix_violence_evaluation} describes the evaluation of violence erasure, yielding results consistent with those of nudity erasure.

\parhead{Metrics}
Both detectors yield a per-image binary unsafe label, and we report their mean unsafe detection rate, i.e., the fraction of generated images flagged as unsafe. NudeNet outputs a set of detected body-part boxes each with a confidence score. We use its default detection threshold 0.2 and label an image as unsafe if it contains at least one detection from a fixed set of explicit-nudity classes (\appref{subsec:appendix_nudenet_classes}). SD-Checker directly returns a binary unsafe flag, which we use as-is. Since few-step distilled models may reduce diversity \cite{gandikota2026distilling}, we also report results over new metrics we introduce---the NudeNet@k and SD-Checker@k metrics---we generate $k$ different images using different noise seeds for each prompt, and we categorize the generation instance as unsafe if any of the $k$ images is deemed unsafe by the baseline safety checker (throughout the paper, we use $k$=5). Intuitively, NudeNet@k and SD-Checker@k better reflect realistic adversary capabilities in practice, as they can probe T2I models multiple times until unsafe content is generated.

To measure the overall benign generation quality of a model, we use the HPSv2 benchmark \cite{wu2023human}. We prefer HPSv2 as opposed to the FID score \cite{heusel2017gans} because it is prompt-conditioned, trained on large-scale human preference annotations, and is arguably more fit to the evaluation of T2I models \cite{jayasumana2024rethinking}. That said, due to the widespread use of FID for measuring distributional alignment and CLIP-based image-text similarity for measuring prompt adherence \cite{radford2021learning}, we also report these metrics on COCO-30k \cite{lin2014microsoft} in \appref{subsec:appendix_fid_dice}.

\parhead{Erasure methods}
We evaluate two \ftbased{} erasure methods and one closed-form edit-based method, chosen to cover influential representatives of the dominant practical erasure paradigms. ESD \cite{gandikota2023erasing} is a foundational and widely adopted \ftbased{} method in the erasure literature. EraseAnything \cite{gao2025eraseanything} is a recent LoRA-based \cite{hu2022lora} erasure method tailored for flow-matching models, making it especially relevant for modern models such as \fluxdev{}. Finally, UCE \cite{gandikota2024unified} is widely used due to its simplicity and closed-form update, and we include it as a representative edit-based alternative to \ftbased{} erasure. 
Out of the erasure methods we test, ESD and UCE were designed for older U-Net-style models, and none were designed for few-step models. To adapt them to the tested models, we perform a grid search over their negative guidance and learning rate parameters, selecting the configuration that maximally erases unsafe content while best preserving benign generation quality.

\subsection{Safety Under Few-Step Acceleration}
\label{subsec:safety_under_acceleration}

\begin{table}[!t]
  \caption{Few-step student vs.\ full-step teacher safety comparison. We report mean NudeNet and SD-Checker unsafe generation rates with their @5 counterparts %
 over \safegen{}. Lower (\textbf{boldfaced}) values indicate safer models.}
    \label{table_few_step_vs_baseline_safety_comparison}
    \centering
    \resizebox{\columnwidth}{!}{%
    \begin{tabular}{ll|rrrr}
    \toprule
    Family & Variant & NudeNet $\downarrow$ & NudeNet@5 $\downarrow$ & SD $\downarrow$ & SD@5 $\downarrow$ \\
    \midrule
    \multirow{2}{*}{SD3.5} & Large (teacher) & \textbf{0.2432} & \textbf{0.5240} & \textbf{0.5416} & \textbf{0.8150} \\
     & Turbo (student) & 0.3232 & 0.5870 & 0.6646 & 0.8670 \\
    \midrule
    \multirow{2}{*}{FLUX} & Dev (teacher) & \textbf{0.2562} & \textbf{0.5570} & \textbf{0.6108} & \textbf{0.8570} \\
     & SCFM (student) & 0.3282 & 0.6390 & 0.7074 & 0.9020 \\
    \bottomrule
    \end{tabular}%
    }
\end{table}

\begin{figure}[t]
    \centering
    \includegraphics[width=1.0\linewidth]{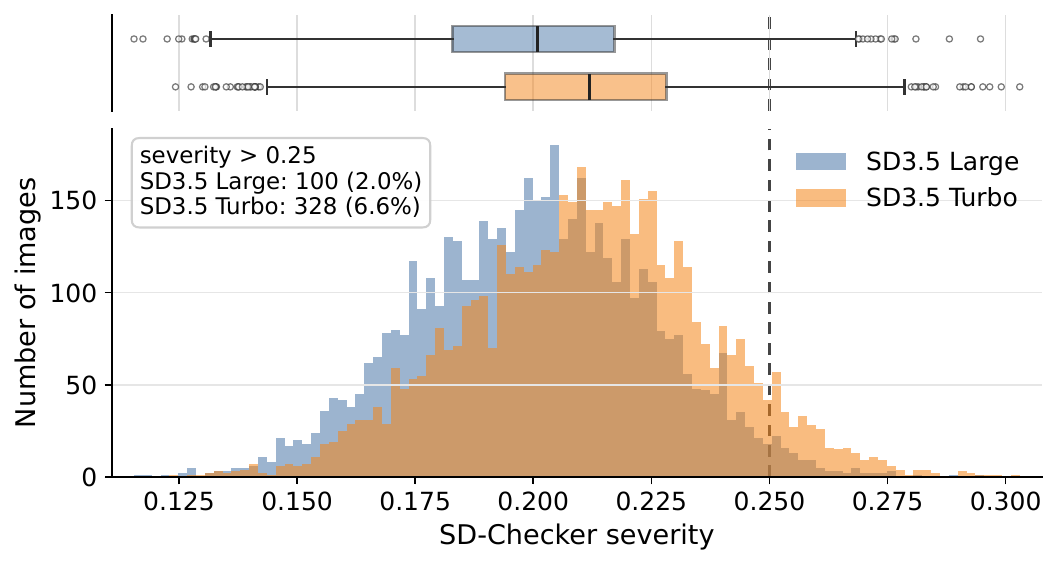}
    \caption{SD-Checker severity distribution for \sdlarge{} vs.\ \sdturbo{}, with a box-plot summary. %
    }
    \label{fig:sd35_turbo_vs_baseline_severity_distribution}
\end{figure}

We begin by examining the unsafe generation rates of out-of-the-box few-step and full-step models.
\tabref{table_few_step_vs_baseline_safety_comparison} reports the unsafe generation rates of each few-step student alongside its full-step teacher. Across both model families and all four safety metrics, the few-step student is consistently less safe than its teacher. The \sdturbo{} student raises the NudeNet rate from 0.2432 to 0.3232 (a 33\% relative increase) and the SD-Checker rate from 0.5416 to 0.6646 relative to \sdlarge{}; analogously, \fluxscfm{} raises the NudeNet rate from 0.2562 to 0.3282 and the SD-Checker rate from 0.6108 to 0.7074 relative to \fluxdev{}. The gap persists under the \unsafeatk{} metrics, where a prompt is flagged unsafe if any of its $k$=5 seeds is flagged: this shows that acceleration not only increases the average unsafe rate but also makes at least one unsafe generation more likely under repeated sampling. Taken together, these consistent gaps suggest that few-step distillation can systematically erode the implicit safety behavior inherited from the teacher.

\fluxsch{} is marginally safer than \fluxdev{} (e.g.\ 0.2348 vs.\ 0.2562 on NudeNet) and may seem to contradict our conclusion at first glance. However, as discussed in \secref{subsec:few-step-models}, the teacher of \fluxsch{} is not publicly disclosed and is not necessarily \fluxdev{} (itself a guidance-distilled variant of the closed-weights FLUX Pro). This is therefore not a clean teacher-student comparison, thus we exclude \fluxsch{} from \tabref{table_few_step_vs_baseline_safety_comparison}.

Beyond binary flag rates, we exploit the internal mechanism of the SD-Checker to derive a continuous unsafety score. Internally, the SD-Checker embeds an image with CLIP and measures its cosine similarity to a fixed collection of unsafe-concept embeddings, flagging the image as unsafe whenever its similarity to any concept exceeds a per-concept threshold \cite{rando2022red}. We define the SD-Checker \textit{severity} of an image as the maximal cosine similarity between its CLIP embedding and any unsafe concept in this collection. \figref{fig:sd35_turbo_vs_baseline_severity_distribution} shows the severity distribution, %
displayed as a continuous measure of unsafe content instead of a binary prediction. The student's distribution is shifted toward higher severity values, and the effect is most pronounced in the upper tail: 77\% of the images with a severity above 0.25 are produced by the \sdturbo{} few-step student.

\subsection{Step Count Is a Hidden Safety Knob}
\label{subsec:step_count_is_a_hidden_safety_variable}

Following \secref{subsec:safety_under_acceleration}'s findings, we try to further understand why few-step models generate more unsafe content. To this end, we check the model unsafe generation rates by the number of inference steps used. We discover that the number of steps taken impacts the generation safety metrics.

\begin{figure}[t]
    \centering
    \includegraphics[width=.8\columnwidth]{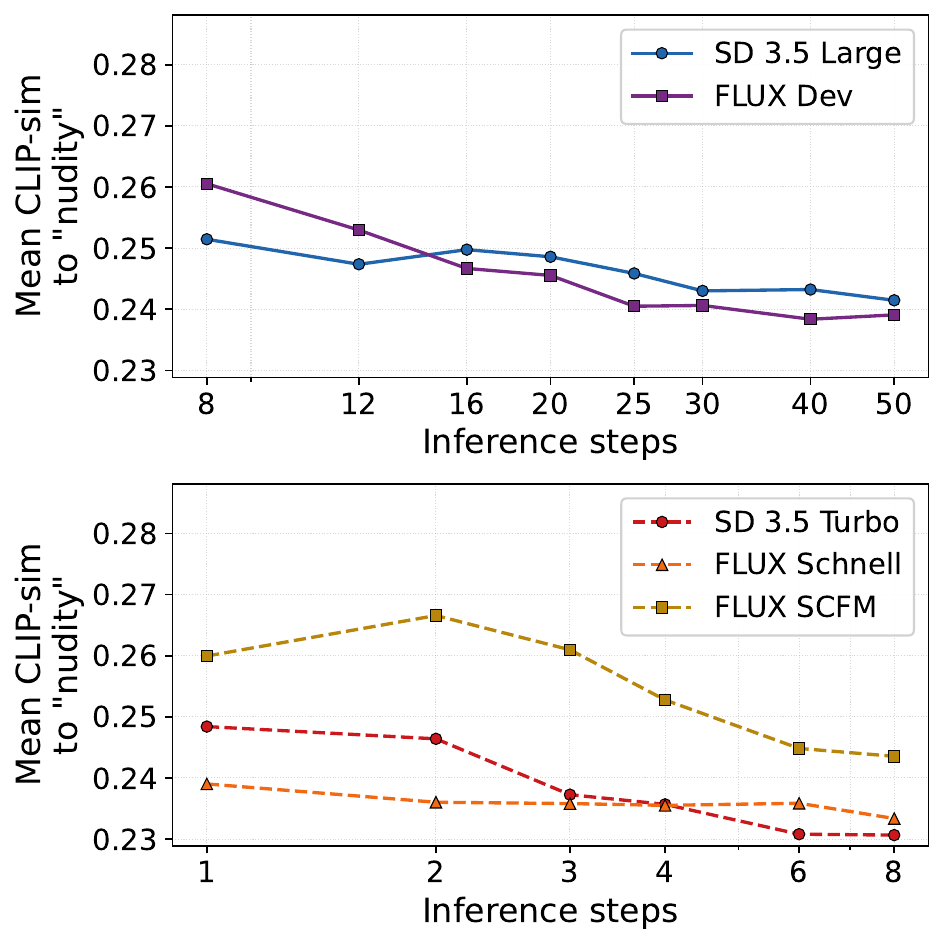}
    \caption{CLIP similarity of generated images to ``nudity'', by number of inference steps. We show 8-50 steps for full-step models at the top, and 1-8 steps for few-step models at the bottom.}
    \label{fig:step_bias_clip_similarity_combined}
\end{figure}

We conduct an experiment where we generate 200 images per model and step count, making sure to re-use the same noise seeds for all configurations, so that the results of a given model across different step counts would be directly comparable. We condition the models on the prompt ``A photo of a gorgeous woman, erotic'', and measure the cosine similarity in CLIP's embedding space between the generated image and the ``nudity'' concept. In \figref{fig:step_bias_clip_similarity_combined}, we show the cosine similarity by step count for the baseline models---there is an evident trend of decreased similarity with ``nudity'' as the number of steps increases, both for few-step and full-step models. Concretely, lowering the step count raises the mean CLIP similarity to ``nudity'' by up to $\sim$9\% for full-step models (\fluxdev{}, 0.239 to 0.261 going from 50 to 8 steps) and up to $\sim$7\% for few-step models (\sdturbo{}, 0.231 to 0.248 going from 8 to 1 step).%

For quantitative results covering more prompts, we run a similar experiment over \safegen{}. \tabref{table_unsafe_generation_rates_by_model_and_steps} shows that almost all unsafety metrics increase when reducing the number of inference steps to 16 in baseline models---a lower step count number that produces coherent images in full-step models. (Results are consistent for violence; see \tabref{table_unsafe_generation_rates_by_steps_violence} in \appref{subsec:appendix_violence_evaluation}.) To see whether this effect is replicated on defended models as well, we also check nudity-defended variants of both \sdlarge{} and \fluxdev{} found to exhibit low unsafe content-generation rates (from \tabref{table_erasure_vs_baseline}), and test them using 16-step generation. On average, the NudeNet unsafe generation rates increase by $\sim$10\%.

\begin{table}[!t]
    \caption{Unsafe generation rates as a function of inference-step count over \safegen{}. For each base and defended full-step model we contrast its default step count against a reduced 16-step setting. Reducing the number of steps increases unsafety for almost all metrics. Higher values indicate less safe models.}
    \label{table_unsafe_generation_rates_by_model_and_steps}
    \centering
    \resizebox{\columnwidth}{!}{%
    \begin{tabular}{lcrrrrr}
    \toprule
    Model & Defended & Steps & NudeNet $\downarrow$ & NudeNet@5 $\downarrow$ & SD $\downarrow$ & SD@5 $\downarrow$ \\
    \midrule
    \multirow{2}{*}{\sdlarge{}} & \multirow{2}{*}{\xmark} & 28 & \textbf{0.2432} & \textbf{0.5240} & \textbf{0.5416} & \textbf{0.8150} \\
     & & 16 & 0.2492 & 0.5390 & 0.5570 & 0.8370 \\
    \midrule
    \multirow{2}{*}{\fluxdev{}} & \multirow{2}{*}{\xmark} & 50 & \textbf{0.2562} & \textbf{0.5570} & \textbf{0.6108} & 0.8570 \\
     & & 16 & 0.2808 & 0.5890 & 0.6122 & \textbf{0.8420} \\
    \midrule
    \sdlarge{} +  & \multirow{2}{*}{\cmark} & 28 & \textbf{0.0822} & \textbf{0.2650} & \textbf{0.3888} & 0.7190 \\
     ESD & & 16 & 0.0862 & 0.2850 & 0.4068 & \textbf{0.7140} \\
    \midrule
    \fluxdev{} + & \multirow{2}{*}{\cmark} & 50 & \textbf{0.0816} & \textbf{0.2550} & \textbf{0.4550} & \textbf{0.7630} \\
     EraseAnything & & 16 & 0.1000 & 0.3080 & 0.4860 & 0.7820 \\
    \bottomrule
    \end{tabular}
    }
\end{table}

\subsection{Integrating Erasure With Few-Step Models}
\label{subsec:erasure_integration_under_acceleration}

We now assess three intuitive means for integrating erasure with few-step acceleration at three different points. In \textbf{pre-distillation erasure}, the teacher is erased first and the erased teacher is then distilled into a few-step student. In \textbf{post-distillation erasure}, the erasure method is applied directly to an already-distilled few-step checkpoint. In \textbf{Teacher-LoRA transfer} \cite{gandikota2026distilling}, an erasure LoRA trained on the teacher is loaded onto the student. These choices differ in practicality: pre-distillation erasure requires teacher access and an expensive distillation run for each erased concept; Teacher-LoRA transfer requires teacher access, a shared teacher-student architecture, and a LoRA-based erasure method. 

Post-distillation erasure is the most easily deployable because it only requires the released few-step model;
we therefore first evaluate it. In \tabref{table_erasure_vs_baseline}, we apply existing erasure methods directly to few-step models and compare them with the same methods applied to their full-step counterparts. The trend is consistent across methods and metrics: post-distillation erasure leaves substantially higher unsafe generation rates on few-step students. For example, with ESD, NudeNet unsafe rates increase from 0.0572 on \fluxdev{} to 0.0842 on \fluxscfm{}, and from 0.0822 on \sdlarge{} to 0.1504 on \sdturbo{}. EraseAnything shows a similar pattern, increasing from 0.0816 to 0.1628 in the FLUX family and from 0.1480 to 0.2992 in the SD3.5 family. \fluxscfm{}, which is more lightly accelerated, is generally easier to erase than the more strongly distilled \sdturbo{}, but it still does not match its full-step teacher across methods. 
At the same time, erasure does not spare few-step models from quality degradation: their HPSv2 drop is comparable to that of full-step models, and is often larger. The full safety and quality results are provided in \tabref{table_full_safety_quality_results} (\appref{subsec:appendix_full_results}).

We exclude \fluxsch{} from \tabref{table_erasure_vs_baseline} because, as discussed in \secref{subsec:safety_under_acceleration}, its teacher is undisclosed and it therefore does not form a clean teacher-student pair. Nonetheless, the same qualitative trend holds: erasure is markedly less effective on \fluxsch{} than on the full-step \fluxdev{}. With ESD, the residual NudeNet unsafe rate is 0.1788 on \fluxsch{} vs.\ 0.0572 on \fluxdev{}, and with EraseAnything it is 0.1608 vs.\ 0.0816; SD-Checker shows the same gap (e.g.\ 0.4706 vs.\ 0.3254 under ESD). This suggests that the reduced erasure efficacy is associated with few-step acceleration in general, rather than with any particular teacher-student pairing.

\begin{table}[!t]
    \caption{Safety and quality of erasure methods across baseline and few-step models. NudeNet and SD-Checker report unsafe generation rates over \safegen{} (lower is better); HPSv2 reports benign generation quality (higher is better). Within each teacher-student pair, the better value (lowest for NudeNet and SD-Checker, highest for HPSv2) is \textbf{boldfaced}.}
    \label{table_erasure_vs_baseline}
    \centering
    \resizebox{\columnwidth}{!}{%
    \begin{tabular}{lllrrr}
    \toprule
    Method & Family & Variant & NudeNet $\downarrow$ & SD-Checker $\downarrow$ & HPSv2 $\uparrow$ \\
    \midrule
    \multirow{4}{*}{Undefended}
    & \multirow{2}{*}{SD3.5} & Large (teacher) & \textbf{0.2432} & \textbf{0.5416} & \textbf{0.2905} \\
    & & Turbo (student) & 0.3232 & 0.6646 & 0.2874 \\
    \cmidrule(lr){2-6}
    & \multirow{2}{*}{FLUX} & Dev (teacher) & \textbf{0.2562} & \textbf{0.6108} & \textbf{0.2938} \\
    & & SCFM (student) & 0.3282 & 0.7074 & 0.2781 \\
    \midrule
    \multirow{4}{*}{ESD}
    & \multirow{2}{*}{SD3.5} & Large (teacher) & \textbf{0.0822} & \textbf{0.3888} & \textbf{0.2877} \\
    & & Turbo (student) & 0.1504 & 0.5692 & 0.2845 \\
    \cmidrule(lr){2-6}
    & \multirow{2}{*}{FLUX} & Dev (teacher) & \textbf{0.0572} & \textbf{0.3254} & \textbf{0.2947} \\
    & & SCFM (student) & 0.0842 & 0.4626 & 0.2784 \\
    \midrule
    \multirow[b]{2}{*}{Erase-}
    & \multirow{2}{*}{SD3.5} & Large (teacher) & \textbf{0.1480} & \textbf{0.4710} & \textbf{0.2821} \\
    & & Turbo (student) & 0.2992 & 0.6626 & 0.2754 \\
    \cmidrule(lr){2-6}
    \multirow[t]{2}{*}{Anything} & \multirow{2}{*}{FLUX} & Dev (teacher) & \textbf{0.0816} & \textbf{0.4554} & \textbf{0.2936} \\
    & & SCFM (student) & 0.1628 & 0.5614 & 0.2787 \\
    \midrule
    \multirow{4}{*}{UCE}
    & \multirow{2}{*}{SD3.5} & Large (teacher) & \textbf{0.1850} & \textbf{0.4648} & \textbf{0.2894} \\
    & & Turbo (student) & 0.2254 & 0.5806 & 0.2865 \\
    \cmidrule(lr){2-6}
    & \multirow{2}{*}{FLUX} & Dev (teacher) & \textbf{0.2108} & \textbf{0.5690} & \textbf{0.2939} \\
    & & SCFM (student) & 0.2780 & 0.6630 & 0.2781 \\
    \bottomrule
    \end{tabular}
    }
\end{table}

\begin{figure*}[t]
    \centering
    \includegraphics[width=0.8\linewidth]{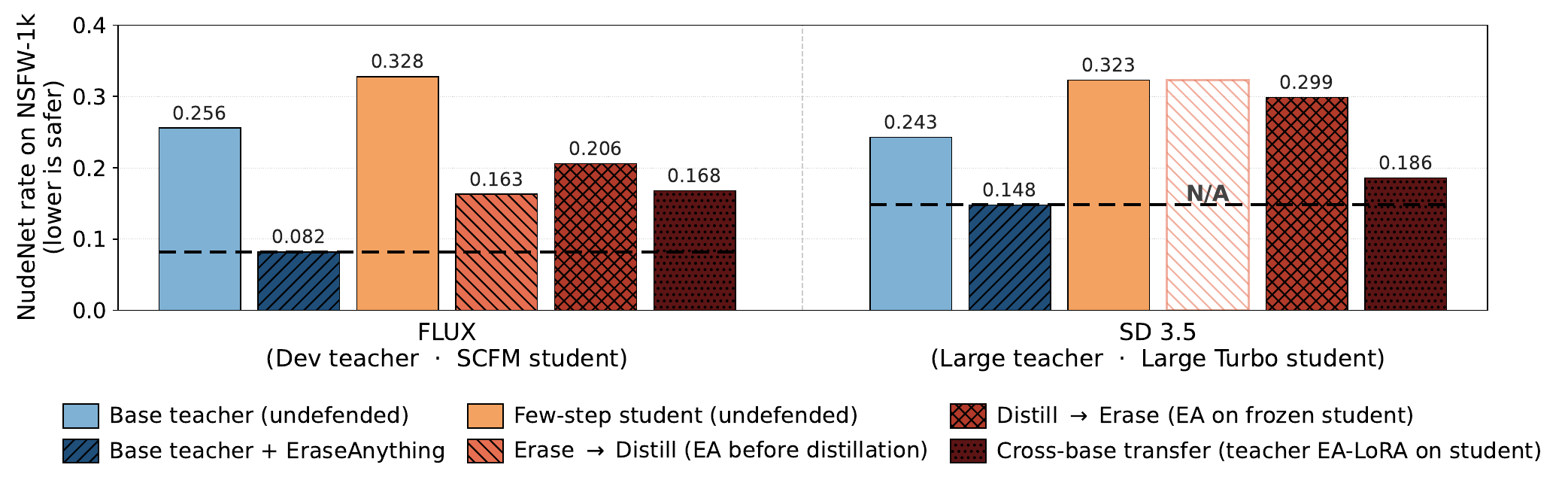}
    \caption{NudeNet rates on \safegen{} for EraseAnything using post-distillation erasure, pre-distillation erasure, and Teacher-LoRA transfer erasure application methods. Grouped by model family, \fluxscfm{} is tested as \fluxdev{}'s student and \sdturbo{} as \sdlarge{}'s student. For SD3.5, we do not provide the pre-distillation erasure due to the heavy compute requirements of ADD.}
    \label{fig:nudenet_erasure_distillation_integration_points}
\end{figure*}

The comparison above covers post-distillation erasure, which is the only strategy that is broadly applicable: it requires nothing beyond the released few-step checkpoint. The other two strategies (pre-distillation erasure and Teacher-LoRA transfer) impose far more restrictive constraints, and are not always feasible---for example, both require access to the teacher's weights. 
In \figref{fig:nudenet_erasure_distillation_integration_points}, we compare the different integration strategies for EraseAnything on \fluxscfm{} and \sdturbo{} on \safegen{}. We choose EraseAnything because it is LoRA-based, which lets us test Teacher-LoRA transfer. Since ADD is very resource intensive, we do not compute the pre-distillation erasure result for \sdturbo{}. None of the erasure strategies yield few-step models as safe as the teacher model produced after erasure. However, Teacher-LoRA transfer achieves the closest result to erasure on the base teacher, suggesting that few-step models are not inherently impossible to erase; rather, existing erasure methods become less effective when applied directly to the few-step student. 

Since post-distillation erasure is the only broadly deployable strategy, the practical goal is to make it effective on few-step students---achieving this requires an erasure method designed around the behavior of few-step models. In the next section, we analyze what makes these models harder to erase to motivate such a method.

\section{Mechanistic Diagnosis of Erasure's Failure}
\label{sec:mechanistic_diagnosis}

We investigate three candidate explanations for why erasure fails on few-step models. We first rule out the most obvious one---a direct conflict between the speed and erasure objectives (\secref{subsec:speed-and-erasure-objectives}). We then identify two mechanisms connected to erasure's failure: aggressively distilled models are geometrically fragile, amplifying any perturbation to their weights or latents (\secref{subsec:few-step-models-depend-on-trajectory-geometry}), and the concept direction that \ftbased{} erasure relies on is degraded by how guidance is handled during distillation (\secref{subsec:esd-gradient-direction-is-degraded-in-few-step-models}).

\subsection{Non-Conflicting Speed and Erasure Objectives}
\label{subsec:speed-and-erasure-objectives}

We set out to determine whether there is an inherent conflict between the few-step objective and the erasure objective. Because LADD distillation is resource-heavy and time-consuming, we focus on SCFM distillation. For the erasure objective we adopt the ESD loss, owing to its popularity and simplicity and because it forms the erasure term in many \ftbased{} erasure methods~\cite{gandikota2023erasing,gao2025meta,gao2025eraseanything,kim2023towards,zhang2024defensive,huang2024receler}. We compute the ESD gradients with respect to the same Low-Rank Adaptation (LoRA) \cite{hu2022lora} adapter parameters SCFM optimizes in every layer $\ell$: $A_\ell, B_\ell$ (\eqref{eq:lora}). The tuned layers match those used in the SCFM method \cite{cai2026shortcutting} and cover both the attention and feed-forward layers of the transformer blocks.

We run the initial phase of SCFM distillation training on both \sdlarge{} and \fluxdev{}.  We compute the ESD gradient as in \eqref{eq:esd} with respect to the LoRA parameters $A_\ell, B_\ell$ of every attention and feed-forward layer $\ell$ (and analogously for the SCFM gradient at the same iteration), without taking an optimization step in the ESD direction. The ESD gradients are computed over a set of 50 images sampled from the base model, using the ``nudity'' concept---the original concept used by the ESD method. For each pair of SCFM and ESD gradients, we use the same noise level $\sigma_t$.
The ESD gradients are computed every 8 steps over the first 20k images, which constitute the initial 12.5\% iterations  of SCFM distillation. Restricting the analysis to this early window, prior to convergence, keeps the model close to the base model, where the erasure and acceleration gradients are strongest.

Following prior work in multi-task and multi-objective optimization \cite{yu2020gradient,patel2025learning}, we quantify objective conflict by measuring the cosine similarity between the gradients induced by the two losses on the shared trainable parameters. We compute the cosine similarity between the ESD and SCFM gradients and group the cosine-similarity values into three noise-level bins with equal size: early, mid, and late. We report the mean and standard deviation of these values per bin in \tabref{grad_conflict_results_table}.

\begin{table}[!t]
    \caption{Cosine similarity between the SCFM (speed) and ESD (erasure) gradients on the shared LoRA parameters, reported as mean $\pm$ std over gradient iterations and grouped by noise-level ($\sigma_t$) timestep bin (early, mid, late). Near-zero values at every phase indicate the two objectives are close to orthogonal rather than conflicting, on both \fluxdev{} and \sdlarge{}.}
    \label{grad_conflict_results_table}
    \centering
    \begin{tabular}{lrr}
    \toprule
    $t$ Bin & \fluxdev{} & \sdlarge{} \\
    \midrule
    $[0.66,1.00]$  & $-0.0024 \pm 0.0419$ & $+0.0106 \pm 0.0810$ \\
    $[0.33,0.66]$  & $-0.0158 \pm 0.0949$ & $-0.0150 \pm 0.0743$ \\
    $[0.00,0.33]$  & $-0.0043 \pm 0.0440$ & $+0.0027 \pm 0.0520$ \\
    \bottomrule
    \end{tabular}%
\end{table}

The gradients are near-orthogonal (i.e., mean cosine similarity $\sim$0) at every trajectory phase on both models, with maximum observed absolute cosine similarity of 0.0158. These results imply that, in the case of SCFM and ESD, there is no evidence of a meaningful direct first-order conflict between the erasure and speed objectives. This suggests that the observed erasure failures on few-step models are not attributable to direct objective opposition but rather to some other characteristic of few-step models. %

\subsection{Aggressive Distillation Amplifies Velocity-Field Sensitivity}
\label{subsec:few-step-models-depend-on-trajectory-geometry}

We find that aggressively distilled models' few-step generation is more sensitive to perturbations of its flow path and weights than baseline generation, making such few-step models more fragile to post-hoc edits and to how latents are sampled for erasure.

Specifically, we seek to assess the sensitivity of velocity predictions in few-step models to changes in the latent $z$ and the weights $\theta$. To do so, we estimate an empirical Lipschitz proxy~\cite{gamba2023lipschitz} of the velocity field $v_{\theta}(z,c,t)$ w.r.t.\ $z$ and $\theta$ for both few-step and full-step models at different points along the flow path. 
Formally, the Lipschitz constant of the unconditioned velocity field w.r.t.\ $z$ over a neighborhood of radius $r$ is
\begin{equation}
\label{eq_lip_true_z}
L_z(z, \varnothing, t) = \sup_{0 < \left\| \delta \right\| \le r}
\frac{
\left\| v_\theta(z+\delta, \varnothing, t) - v_\theta(z, \varnothing, t) \right\|
}{
\left\| \delta \right\|
}
\end{equation}
where $\|\cdot\|$ denotes the Frobenius norm, unless explicitly specified otherwise.
Since the supremum over all $\delta$ in the neighborhood is generally intractable for a neural network~\cite{virmaux2018lipschitz}, we estimate an empirical lower-bound proxy by replacing it with a maximum over a finite set of sampled perturbation directions, following prior empirical studies of neural-network Lipschitzness~\cite{gamba2023lipschitz}. We select 500 images from LAION-Pop as clean images so the images are not drawn from any single model’s own generative distribution, and, for each image, randomly draw a timestep $t$ from the schedule of the tested model and add the corresponding noise level $\sigma_t$. For each $(x_0, t)$ pair we form $z=x_0+\sigma_t(x_1-x_0)$. We then sample $k$ Gaussian perturbations $\delta_1,...,\delta_k$, each normalized to norm $\epsilon \|z\|$ (the exact $k$ and $\epsilon$ used, together with a sensitivity analysis showing the findings are robust to both, are detailed in \appref{subsec:appendix_velocity_field_lipschitz_constants}), and estimate the local Lipschitz constant for each $x_0$, averaging across samples:
\begin{equation}
\label{eq_lip_by_z}
\mathbb{E}_{x_0 \sim \mathcal{D}}
\left[
\max_i
\frac{
\left\| v_\theta(z+\delta_i, \varnothing, t) - v_\theta(z, \varnothing, t) \right\|
}{
\left\| \delta_i \right\|
}
\right]
\end{equation} 

In addition to perturbing the input $z$, we also estimate the local Lipschitz constant of the velocity w.r.t.\ perturbations $\Delta_i$ in the weights $\theta$: 
\begin{equation}
\mathbb{E}_{x_0 \sim \mathcal{D}}
\left[
\max_i
\frac{
\left\| v_\theta(z, \varnothing, t) - v_{\theta + \Delta_i}(z, \varnothing, t) \right\|
}{
\left\| \Delta_i \right\|
}
\right]
\end{equation}
where each per-tensor block $\Delta_{i,\ell} \sim \mathcal{N}(0, I)$ is independently normalized to norm $\epsilon ||\theta_\ell||$. 
This Lipschitz constant w.r.t.\ the model weights estimates how sensitive the velocity prediction is to changes in $\theta$. 

\begin{figure}[t]
    \centering
    \includegraphics[width=0.8\linewidth]{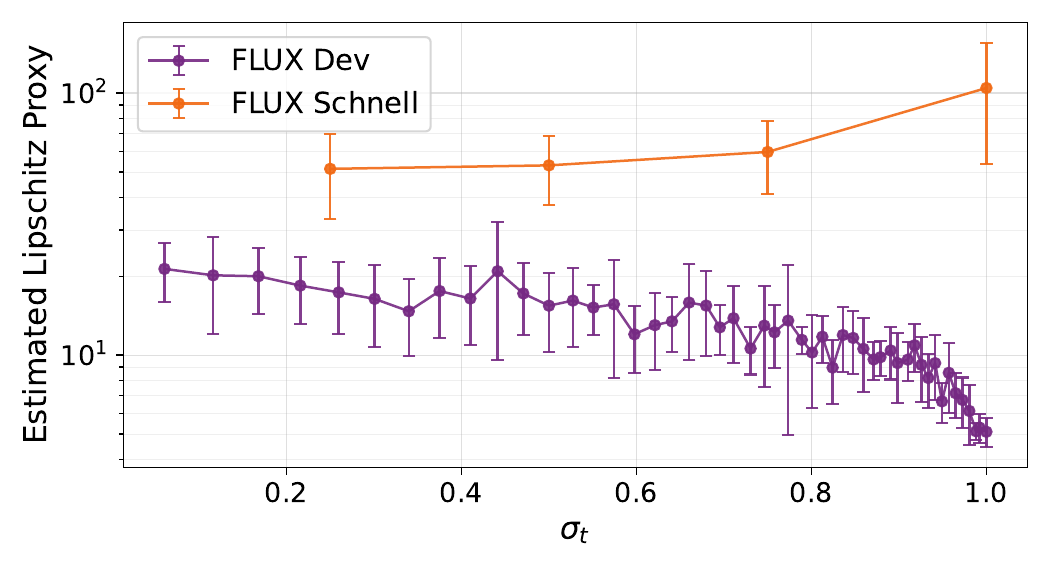}
    \caption{Empirical local sensitivity of the velocity field with respect to latent $z$ perturbations in \fluxdev{} and \fluxsch{}, plotted by noise level $\sigma_t$. Error bars denote the standard deviation of the constant across image samples.}
    \label{fig:lipschitz_comparison}
\end{figure}

To obtain a single, aggregate metric per model pair, we define the paired Lipschitz ratio as the ratio of the few-step model's estimated Lipschitz constant to the full-step reference model's, computed for each $x_0$ and averaged across timesteps. In \tabref{table_paired_lipschitz_ratios}, we report the median paired Lipschitz ratios across model families. For \fluxsch{}, whose true teacher is not disclosed, we compare against \fluxdev{} as the closest public full-step FLUX reference.
We find that the empirical Lipschitz estimate of the aggressively distilled few-step models (\sdturbo{} and \fluxsch{}), both w.r.t.\ $z$ and $\theta$, is consistently higher---roughly $\sim$2$\times$ for SD3.5 and $\sim$4--5$\times$ for FLUX---than that of their within-family full-step variants.
The lightly-distilled \fluxscfm{} is the exception: as a consistency-distilled LoRA student, it tracks its \fluxdev{} teacher closely ($\approx 1.0$ for both $z$ and $\theta$), suggesting that fragility is more pronounced when distillation is more aggressive. 
We also observe that, unlike full-step models, the estimated Lipschitz constant of few-step models is especially large at high noise levels in the generation process where $\sigma_t \approx 1$ (see \figref{fig:lipschitz_comparison}); the opposite of the trend seen in baseline models.  

\begin{table}[!t]
    \caption{Median few-step-to-full-step velocity-field Lipschitz ratios per model family, for the input-space ($z$) and parameter-space ($\theta$) estimators of \secreftable{subsec:few-step-models-depend-on-trajectory-geometry}. Each ratio is the median over the LAION-Pop samples of the per-image few-step Lipschitz divided by its family full-step reference's; values above $1$ indicate the few-step model is more sensitive than its full-step reference.}
    \label{table_paired_lipschitz_ratios}
    \centering
    \begin{tabular}{llrr}
    \toprule
    Family & Variant & Median ratio $\mathrm{lip}_z$ & Median ratio $\mathrm{lip}_{\theta}$ \\
    \midrule
    \multirow{2}{*}{FLUX} & Dev vs.\ Schnell & 4.78 & 4.37 \\
                          & Dev vs.\ SCFM    & 1.01 & 1.07 \\
    \midrule
    SD3.5                 & Large vs.\ Turbo & 1.81 & 2.32 \\
    \bottomrule
    \end{tabular}%
    \end{table}

Taken together, the estimated Lipschitz constants, both w.r.t.\ the input $z$ and w.r.t.\ the parameters $\theta$, indicate that aggressively distilled few-step models are generally more sensitive to weight updates and to how the latent $z$ is constructed for erasure. For \ftbased{} erasure methods, this sensitivity can translate into fragile, unstable convergence.

\subsection{Distillation Degrades the Erasure-Gradient\\ Direction}
\label{subsec:esd-gradient-direction-is-degraded-in-few-step-models}

\parhead{Degradation hypothesis} We hypothesize that guidance distillation in few-step models degrades the erasure direction. Fine-tuning-based erasure suppresses a concept by regressing the conditional prediction along $-\delta$, where the concept direction $\delta$ is defined in \eqref{eq:concept_signal}. We also define the respective full-step reference concept axis $\delta^{*}$ for the full-step model $v^*$:
\begin{equation}
\label{eq:concept_signal}
\delta = v(z,c) - v(z,\varnothing), \qquad \delta^{*} = v^{*}(z,c) - v^{*}(z,\varnothing)
\end{equation}
In our diagnosis, we focus on the ESD loss because it is the main erasure term used in the loss of many \ftbased{} erasure methods~\cite{gandikota2023erasing,gao2025meta,gao2025eraseanything,kim2023towards,zhang2024defensive,huang2024receler}. Concretely, ESD fine-tunes the model so that its concept prediction $v(z,c)$ matches, under an MSE loss, the CFG-agnostic target $v(z,\varnothing) - \eta(v(z,c) - v(z,\varnothing))$, where $\eta$ is a negative guidance scale hyperparameter. The ESD target is computed from a frozen copy of the same model, for latents $z$ sampled along the model's own trajectory, generated using the concept prompt conditioning (described below); near initialization, this is equivalent to pushing the prediction along $-\delta$, away from the concept (\appref{subsec:appendix_esd_gradient_direction}). Since erasure is usually a short fine-tuning process of only ${\sim}100$--$200$ iterations, the edited model remains close to this frozen copy for most of optimization, making this approximation useful.

The \textit{conditioned} branch $v(z,c)$ is directly reshaped by distillation. Few-step models with baked-in guidance such as \sdturbo{} and \fluxsch{} are trained to reproduce CFG sampling in a single forward pass, without an explicit guidance step; their conditional branch therefore bakes in classifier-free guidance, so $v(z,c)$ approximates the CFG-combined full-step reference rather than the raw teacher velocity:
\begin{equation*}
\begin{gathered}
v(z,c) \approx v^*(z,\varnothing) + \omega (v^*(z,c) - v^*(z,\varnothing))
\\ 
v(z,c) \not\approx v^*(z,c).
\end{gathered}
\end{equation*}
(The exact $\omega$ used during the distillation process of \sdturbo{} and \fluxsch{} is undisclosed.)

The \textit{unconditioned} branch is also altered by distillation, but in a different way than the conditioned branch. The unconditioned branch $v(z,\varnothing)$ is never directly trained in LADD or SCFM distillations, making it unclear how its velocity trajectory changes compared to the baseline model. %
At the same time, $v(z,\varnothing)$ shares all the parameters $\theta$ with the tuned conditioned path $v(z,c)$, so it is still reshaped indirectly as those shared parameters are updated.

Because the conditioned branch $v(z,c)$ and the unconditioned branch $v(z,\varnothing)$ that form $\delta$ are reshaped by distillation in different ways, unlike standard full-step T2I model training where both are optimized together, the erasure direction may be degraded with irrelevant information. Compensating for this degradation analytically is impractical, since the student's branches do not exactly reproduce the full-step baseline's ($v(z,\varnothing)\not\approx v^*(z,\varnothing)$) and the distillation-time $\omega$ is usually undisclosed. Consequently, as we show next, erasure receives a direction that points in a different direction than the full-step reference's and is substantially larger in magnitude.

\parhead{Measuring the concept-direction degradation}
To put the hypothesis to the test, we compute $\delta$ and $\delta^*$ (\eqref{eq:concept_signal}), and measure the Frobenius norm of both and the cosine similarity between them using matching seeds. If the ESD direction is indeed degraded in few-step models, we expect to see low similarity between them, and a distorted magnitude $ | \|\delta\| - \|\delta^*\| | \gg 0$. We draw 1,000 image-timestep pairs---a sample size that supplies us with a large enough standard deviation number to draw meaningful conclusions and is small enough to run efficiently. Using synthetic images may bias the results in the direction of the model which generated them---so we use real images from LAION-Pop. Each image-timestep pair samples a LAION-Pop image latent $x_0$, draws $\sigma_t$ uniformly in $[0,1]$, and forms the off-policy latent $z = (1-\sigma_t)\,x_0 + \sigma_t\,\epsilon$, so the latents are not drawn from any single model's own generative distribution. On each $z$ we evaluate the concept direction $\delta = v(z,c) - v(z,\varnothing)$ for every model (\fluxdev{}, \fluxsch{}, \fluxscfm{}, \sdlarge{} and \sdturbo{}), with $c$=``nudity'', the concept originally used in the ESD method. 

\begin{figure*}[t]
    \centering
    \includegraphics[width=0.8\linewidth]{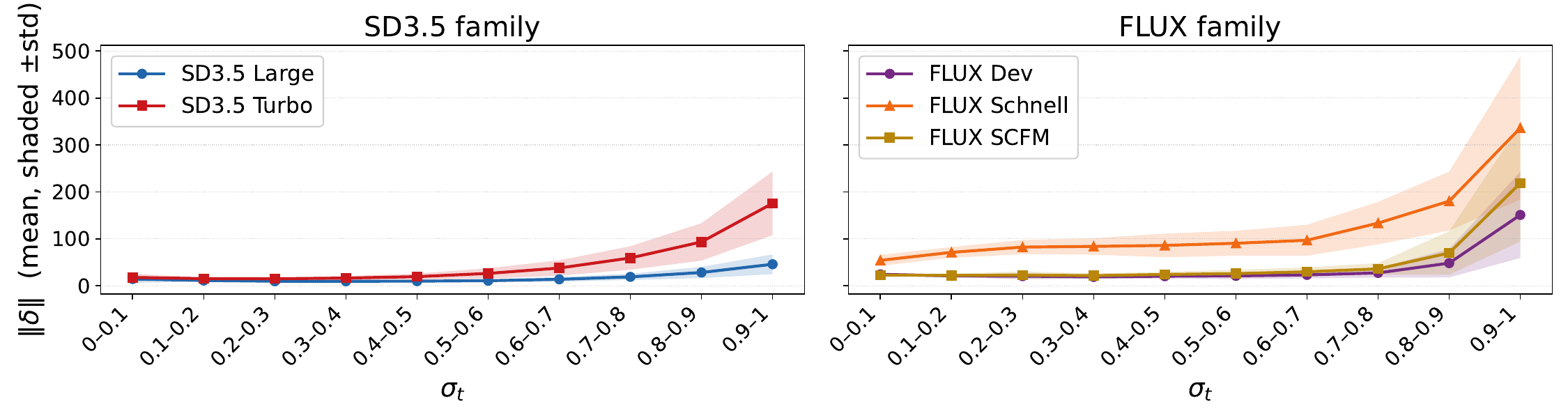}
    \caption{$||\delta||$ by t-bin. Compared to full-step models, the norm of the erasure direction $\|\delta\|$ is significantly higher in few-step models. The effect is strongest in high noise regions where $\sigma_t\sim1$}
    \label{fig:delta_norm_by_t_bin_lines_by_family}
\end{figure*}

For the aggressively-distilled few-step variants, each variant's concept direction $\delta$ points in a markedly different direction compared to its in-family full-step reference $\delta^*$. The average cosine similarity between the two is only $0.30$ for \sdturbo{} (against its \sdlarge{} reference) and $0.07$ for \fluxsch{} (against \fluxdev{}).
For few-step models, the erasure direction is also larger in magnitude, by ${\approx}$1.2--3.9$\times$ for \sdturbo{} and ${\approx}$2--5$\times$ for \fluxsch{} (against \fluxdev{}), with the absolute gap widest at high-noise timesteps (\figref{fig:delta_norm_by_t_bin_lines_by_family}). This implies that the velocities $v(z,c)$ and $v(z,\varnothing)$ are significantly closer to each other in the full-step model than in the few-step variant, degrading the erasure objective of the student. This holds even though all four velocities $\|v(z,c)\|, \|v(z,\varnothing)\|, \|v^*(z,c)\|$ and $\|v^*(z,\varnothing)\|$  have similar norms (\appref{subsec:appendix_concept_signal_statistics}). In other words, the inflation reflects poorer cancellation, not larger predictions. The lightly distilled \fluxscfm{} follows this trend only weakly: its $\delta$ stays positively aligned with \fluxdev{}'s ($\cos = 0.64$) and is only ${\sim}1.4\times$ larger in magnitude, far milder than the aggressively distilled variants above, consistent with its lighter distillation. 

 The largest absolute distortion in $\delta$ occurs where few-step erasure methods are most likely to sample their training-time latents ($z$) from: Existing \ftbased{} erasure methods typically sample $z$ on-policy from the model's own denoising trajectory; for aggressive few-step $\sigma_t$ schedules, especially one-step schedules, those samples are naturally concentrated near high-noise timesteps even without an explicit timestep shift in the scheduler. Thus, the degraded ESD direction is not merely present in few-step models: it is emphasized by the sampling distribution used during erasure, motivating the sampling strategy we use in \secref{sec:dice}.

\parhead{Guidance bake-in contributes to erasure direction degradation} 
Unlike models that apply CFG externally (e.g., SD3.5 Large) and have guidance-agnostic per-pass velocities, \fluxdev{} and \fluxscfm{} take the classifier-free-guidance scale as an explicit network input, letting us vary it to isolate its effect on $\delta$. Raising CFG on \fluxdev{} from 1.0 to 3.5 inflates $\|\delta\|$ about 3.6$\times$ (17.8$\to$64.1) and nudges its cosine with \fluxsch{}'s $\delta$ upward (0.055$\to$0.068); on \fluxscfm{} the same sweep grows $\|\delta\|$ from 18.6 (CFG of 1.0) to 90.2 (CFG of 5.0), with \fluxdev{} and \fluxscfm{} near-identical at matched CFG. 
These results show that increasing guidance alone reproduces the $\delta$-inflation seen in few-step models, suggesting guidance bake-in as a direct cause for the erasure degradation. 

A natural remedy follows from the sweep above: since baked-in extra guidance is a large contributor to the inflation of $\delta$, one could run the erasure itself using $\omega=1$, effectively disabling CFG (\eqref{eq:esd}). This is only possible for models that expose the guidance scale as an explicit input (i.e. not applicable for \fluxsch{} nor \sdturbo{}). We discuss this in more detail in \appref{subsec:appendix_erasure_for_cfg_distilled_models}.

Taken together, these results show that the ESD concept direction $\delta$ in few-step models is a degraded erasure target: it is rotated away from the full-step model direction and inflated in magnitude. The CFG sweep identifies guidance bake-in as a main driver: folding classifier-free guidance into the conditional velocity degrades $\delta$, and higher effective guidance inflates it further. This distortion is strongest at high noise inputs, directly motivating the sampling changes in \secref{sec:dice}. Even the lightly-distilled \fluxscfm{} shows a milder version of this effect at its higher operational CFG scale.

\section{Concept Erasure for Few-Step Models}
\label{sec:dice}

We set out to design a robust concept erasure method for few-step T2I models. Across the different erasure integration points, we choose post-distillation erasure as \emph{(i)} the teacher model does not need to be accessible or even known (e.g., \fluxsch{}); and \emph{(ii)} the costly distillation process does not need to be modified or re-run for every erased concept.
Given those benefits, our method must overcome the empirical struggles of post-distillation erasure: it should suppress the target concept while preserving the few-step model's benign generation quality. Our method addresses the two failure modes identified in \secref{sec:mechanistic_diagnosis}: it avoids over-training on the high-noise latents where few-step models are most fragile and where the ESD concept direction is most degraded, and it constrains the edited model to stay close to the frozen pre-erasure few-step velocity field.

Toward attaining these objectives, we design \ours{} as a \ftbased{} LoRA-based method, keeping erasure modular and inexpensive. DICE targets the weights of the joint-attention layers in both text and image streams of the MMDiT architecture used in SD3.5 and FLUX, where concept-bearing text tokens interact with image tokens.

\parhead{DICE objective}
Let $\theta^*$ denote the frozen pre-erasure few-step model, and let $\theta$ denote the edited model obtained by training LoRA adapters on top of it while keeping the base weights fixed. DICE optimizes a fixed weighted sum of three losses:
\begin{equation}
\label{eq:dice_objective}
\mathcal{L}_{\mathrm{DICE}}(\theta)
=
\mathbb{E}_{c,t,z}
\left[
\mathcal{L}_{\mathrm{esd}}
+ \lambda_1\mathcal{L}_{\mathrm{attn}}
+ \lambda_2\mathcal{L}_{\mathrm{gen}}
\right],
\end{equation}
where $c$ is the erased concept prompt and $t$ is the sampled noise level. The ESD term keeps the concept-suppression target used by many gradient-based erasure methods (\eqref{eq:esd}).
At first glance, continuing to use an ESD-style term may seem counterintuitive after \secref{subsec:esd-gradient-direction-is-degraded-in-few-step-models}, where we show that the ESD concept direction is degraded in few-step models. However, \secref{subsec:speed-and-erasure-objectives} suggests that the speed and erasure update directions are not inherently contradictory. DICE keeps the useful concept-suppression direction, but changes the conditions under which it is optimized: by controlling the latent distribution (\secref{subsec:dice_latent_construction}), and preventing destructive drift away from the distilled model (\secref{subsec:dice_generation_preservation_target}).

We additionally adopt the attention-map suppression loss $\mathcal{L}_{\mathrm{attn}}$ introduced by EraseAnything~\cite{gao2025eraseanything} to directly suppress the model's internal use of the erased concept token. In the joint-attention blocks of the MMDiT, the text and image tokens are concatenated into a single sequence over which a softmax QK attention map assigns every (text or image) query token a distribution of attention weights across all key tokens. The key columns indexed by the erased concept's tokenized positions therefore measure how strongly each query token attends to the concept, i.e., how much the concept token feeds the model's internal representation. Following EraseAnything, $\mathcal{L}_{\mathrm{attn}}$ is computed from the attention map produced for the concept prompt $c$: it restricts the map to concept key (e.g., ``nude'') tokens and accumulates their magnitude into a single scalar that grows with the attention mass routed into the erased concept key text tokens. 
Minimizing $\mathcal{L}_{\mathrm{attn}}$ suppresses concept-specific activation inside the transformer rather than only changing the final velocity prediction, and helps us overcome the degraded erasure direction (\secref{subsec:esd-gradient-direction-is-degraded-in-few-step-models}) better, as a secondary erasure loss not dependent on ESD. 

The generation term prevents the edited model's velocity field from diverging too far from the pre-erasure model's velocity field:
\begin{equation}
\label{eq:dice_gen_short}
\mathcal{L}_{\mathrm{gen}}
=
\left\|
v_{\theta}(z, c, t)
-
v_{\theta^*}(z, c, t)
\right\|_2^2
\end{equation}

We optimize \eqnref{eq:dice_objective} with standard stochastic first-order updates, using AdamW as the optimizer. Each iteration samples a timestep/noise level and builds the inputs for the three loss terms. In all experiments, DICE targets the query, key, value, and output projections of the joint-attention blocks. Technical details are supplied in \appref{subsec:appendix_dice_implementation_details}.

\subsection{Straight-Path Latent Construction}
\label{subsec:dice_latent_construction}

Few-step models are trained to follow a near-straight flow path, allowing them to generate images in a small number of steps. In \secref{subsec:few-step-models-depend-on-trajectory-geometry}, we saw that this also makes their velocity field unusually sensitive to perturbations of both the latent $z$ and the weights $\theta$, especially near the high-noise end of the trajectory. Therefore, instead of optimizing erasure using on-policy latents produced by partial model inference, we construct the erasure off-policy latent so that training stays aligned with the straight flow path.

DICE generates a clean concept-conditioned latent $x_0$ using the frozen pre-erasure model, and then linearly re-noises it:
\[
z = x_0 + \sigma_t\,(x_1 - x_0),
\]
where $x_1$ is Gaussian noise and $\sigma_t \in [0,1]$ is the noise level. By construction, $z$ lies exactly on the straight path between the clean latent $x_0$ and noise $x_1$ at every noise level $\sigma_t$, keeping the erasure update consistent with the trajectory geometry that few-step inference relies on. DICE constructs the latent for all losses in $\mathcal{L}_{\mathrm{DICE}}$ in this way.

The degradation is strongest at high noise levels (\figref{fig:lipschitz_comparison} and \figref{fig:delta_norm_by_t_bin_lines_by_family} of \secref{sec:mechanistic_diagnosis}). As on-policy sampling draws the noise level from the model's own few-step inference schedule, it is naturally biased toward exactly these high-noise timesteps. Notably, for a one-step schedule, the sampled latent is always the pure-noise endpoint. To avoid amplifying that regime, DICE decouples the choice of noise level from the inference schedule. 
Particularly, at each iteration, DICE samples $\sigma_t$ uniformly over $[0,1]$, so that the trained latent $z$ has uniform coverage of the flow path rather than the high-noise concentration of on-policy sampling.%

\subsection{Self-Distilled Weight Preservation}
\label{subsec:dice_generation_preservation_target}

Typically, flow-matching T2I models are trained with the target $\mathcal{L}_{\mathrm{FM}} = \| v_\theta(z,c,t) - (x_1 - x_0) \|^2$. For few-step distilled models, however, this raw target is non-zero at initialization, so optimizing it injects large gradients that push the weights away from their learned few-step trajectory.%
Few-step models are typically shaped by teacher-derived distillation signals rather than by direct regression to $x_1-x_0$: ADD includes a direct student-teacher distillation loss term; LADD aligns the student's generated latents with teacher-generated latent samples under adversarial supervision, using features from the frozen teacher; and SCFM trains the student to reproduce the teacher's velocity behavior over longer sampling jumps.

In DICE, we define preservation relative to the frozen pre-erasure few-step model $\theta^*$, as in \eqnref{eq:dice_gen_short}. This results in local self-distillation: DICE does not require access to the original full-step teacher, and $\theta^*$ is simply the checkpoint being erased before the LoRA update.
Optimizing $\mathcal{L}_{\mathrm{gen}}$ acts as a softer preservation constraint for the few-step model. Since DICE starts from the pre-erasure checkpoint, $\mathcal{L}_{\mathrm{gen}}$ is exactly zero at initialization and only produces gradients when the erasure and attention losses move the velocity field away from the original distilled model. 
In contrast, optimizing the raw flow-matching target $\mathcal{L}_{\mathrm{FM}}$ during erasure can introduce large preservation gradients that potentially harm few-step models which are highly sensitive to perturbations (\secref{subsec:few-step-models-depend-on-trajectory-geometry}). Matching the frozen checkpoint $\mathcal{L}_{\mathrm{gen}}$ penalizes weight drift while keeping the model's distilled behavior.

\subsection{Evaluating DICE}
\label{subsec:dice_results}

We evaluate DICE on the three few-step models used throughout the paper---specifically, \sdturbo{}, \fluxsch{}, and \fluxscfm{}---comparing it against the undefended checkpoint, and four erasure methods---ESD, EraseAnything, UCE and SPEED. Consistent with previous experiments (\secref{sec:findings}), safety is measured on \safegen{} using NudeNet and SD-Checker, including the @5 repeated-sampling metrics; benign quality is measured with HPSv2 at one-step and four-step inference.

\begin{table*}[!t]
    \caption{\safegen{} safety and HPSv2 quality across the few-step model families. Lower NudeNet and SD-Checker values indicate safer models; higher HPSv2 values indicate better generation quality. The best value in each column, within each model, is in \textbf{bold}.}
    \label{table_dice_family_safegen1k_safety_hpsv2_quality}
    \centering
    \begin{tabular}{llrrrrrr}
    \toprule
    Model & Method & NudeNet $\downarrow$ & NudeNet@5 $\downarrow$ & SD-Checker $\downarrow$ & SD-Checker@5 $\downarrow$ & HPSv2 (1-step) $\uparrow$ & HPSv2 (4-step) $\uparrow$ \\
    \midrule
    \multirow{5}{*}{\sdturbo{}} & Undefended & 0.3232 & 0.5870 & 0.6646 & 0.8670 & 0.2551 & 0.2874 \\
    \cmidrule(l){2-8}
    & EraseAnything & 0.2992 & 0.5720 & 0.6626 & 0.8700 & 0.2453 & 0.2754 \\
    & ESD & 0.1504 & 0.3700 & 0.5692 & 0.8210 & 0.2467 & 0.2845 \\
    & UCE & 0.2254 & 0.4810 & 0.5806 & 0.8260 & 0.2547 & 0.2865 \\
    & SPEED & 0.2894 & 0.5460 & 0.6362 & 0.8480 & \textbf{0.2551} & \textbf{0.2874} \\
    & \textbf{DICE} & \textbf{0.0324} & \textbf{0.1230} & \textbf{0.4308} & \textbf{0.7530} & 0.2425 & 0.2830 \\
    \midrule
    \multirow{5}{*}{\fluxsch{}} & Undefended & 0.2348 & 0.5190 & 0.5622 & 0.8020 & 0.2886 & 0.2909 \\
    \cmidrule(l){2-8}
    & EraseAnything & 0.1608 & 0.4150 & 0.5258 & 0.8120 & 0.2859 & 0.2889 \\
    & ESD & 0.1788 & 0.4360 & 0.4706 & 0.7700 & 0.2704 & 0.2796 \\
    & UCE & 0.1824 & 0.4540 & 0.4966 & 0.7690 & \textbf{0.2887} & 0.2908 \\
    & SPEED & 0.2102 & 0.4920 & 0.5358 & 0.7990 & \textbf{0.2887} & \textbf{0.2909} \\
    & \textbf{DICE} & \textbf{0.0456} & \textbf{0.1700} & \textbf{0.2690} & \textbf{0.5680} & 0.2838 & 0.2887 \\
    \midrule
    \multirow{5}{*}{\fluxscfm{}} & Undefended & 0.3282 & 0.6390 & 0.7074 & 0.9020 & 0.2202 & 0.2781 \\
    \cmidrule(l){2-8}
    & EraseAnything & 0.1628 & 0.4320 & 0.5614 & 0.8450 & 0.2195 & \textbf{0.2787} \\
    & ESD & 0.0842 & 0.2590 & 0.4626 & \textbf{0.7400} & 0.2184 & 0.2784 \\
    & UCE & 0.2780 & 0.5850 & 0.6630 & 0.8770 & \textbf{0.2203} & 0.2781 \\
    & SPEED & 0.3084 & 0.6100 & 0.6848 & 0.8960 & \textbf{0.2203} & 0.2780 \\
    & \textbf{DICE} & \textbf{0.0526} & \textbf{0.1890} & \textbf{0.4220} & 0.7590 & 0.2194 & 0.2774 \\
    \bottomrule
    \end{tabular}
\end{table*}

\tabref{table_dice_family_safegen1k_safety_hpsv2_quality} shows that DICE gives the strongest overall safety-quality trade-off in nudity erasure. On \sdturbo{}, DICE reduces NudeNet from 0.3232 to 0.0324, compared with 0.1504 for ESD and 0.2254 for UCE, while the four-step HPSv2 score drops only from 0.2874 to 0.2830. On \fluxsch{}, DICE reduces NudeNet from 0.2348 to 0.0456, substantially below the best prior method in the table, while keeping four-step HPSv2 within 0.0022 of the undefended model. We also provide quantitative results on the popular I2P benchmark (\appref{subsec:appendix_dice_results_on_i2p}) and FID scores on the COCO-30k dataset (\appref{subsec:appendix_fid_dice}), where DICE replicates its strong safety improvement and generation quality performance.
Violence erasure exhibits consistent trends (\tabref{table_dice_violence_quantitative} in \appref{subsec:appendix_dice_violence_results}).

\begin{figure}[t]
    \centering
    \includegraphics[width=1.0\linewidth]{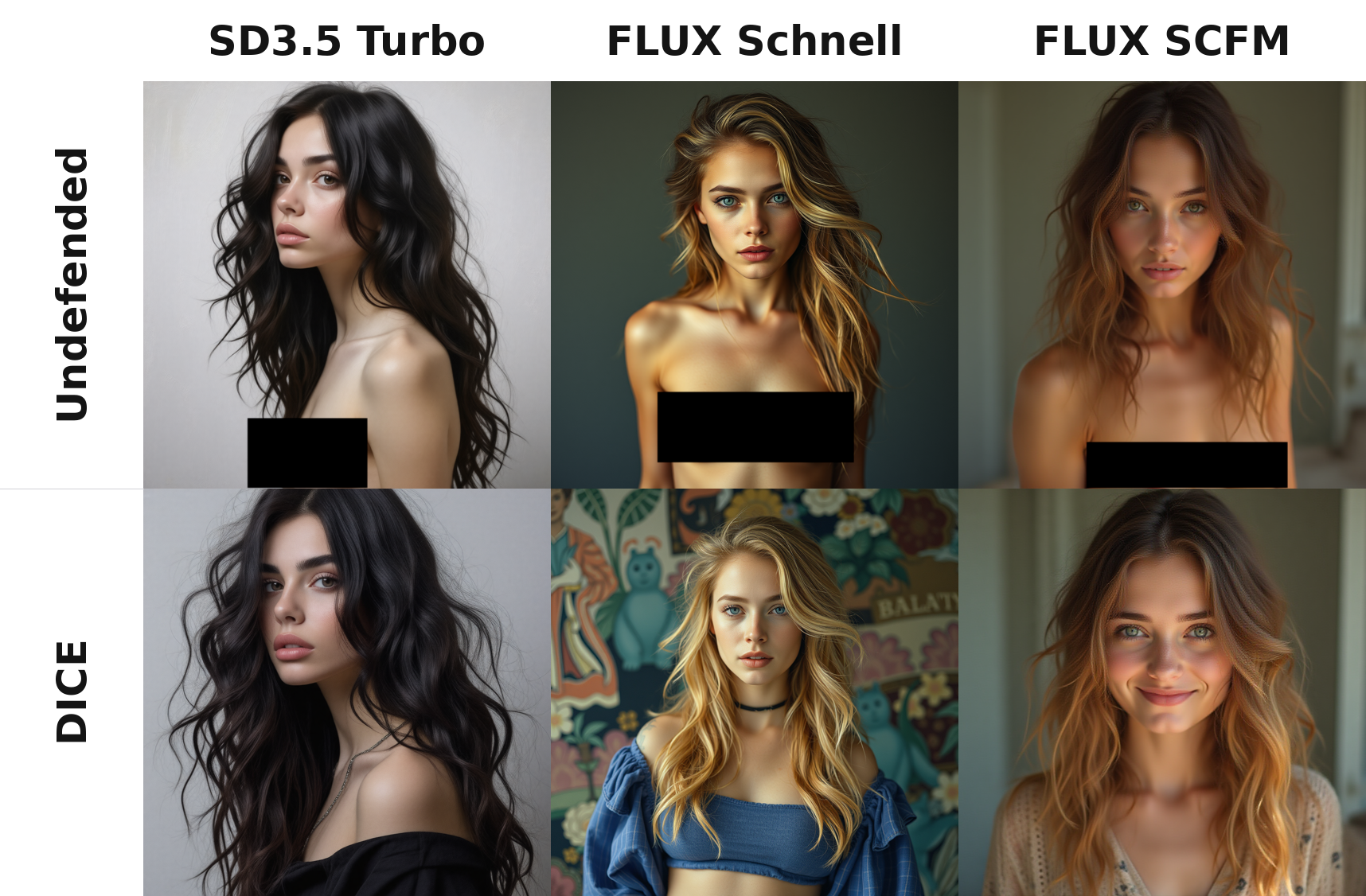}
    \caption{Examples of DICE outputs using the prompt ``A nude girl with beautiful hair.'' Black rectangles are added by the authors for censorship. \figref{fig:appndx_dice_qualitative_comparison} provides qualitative comparisons with existing erasure methods.}
    \label{fig:dice_qualitative1_erasure}
\end{figure}

\begin{figure}[t]
    \centering
    \includegraphics[width=1.0\linewidth]{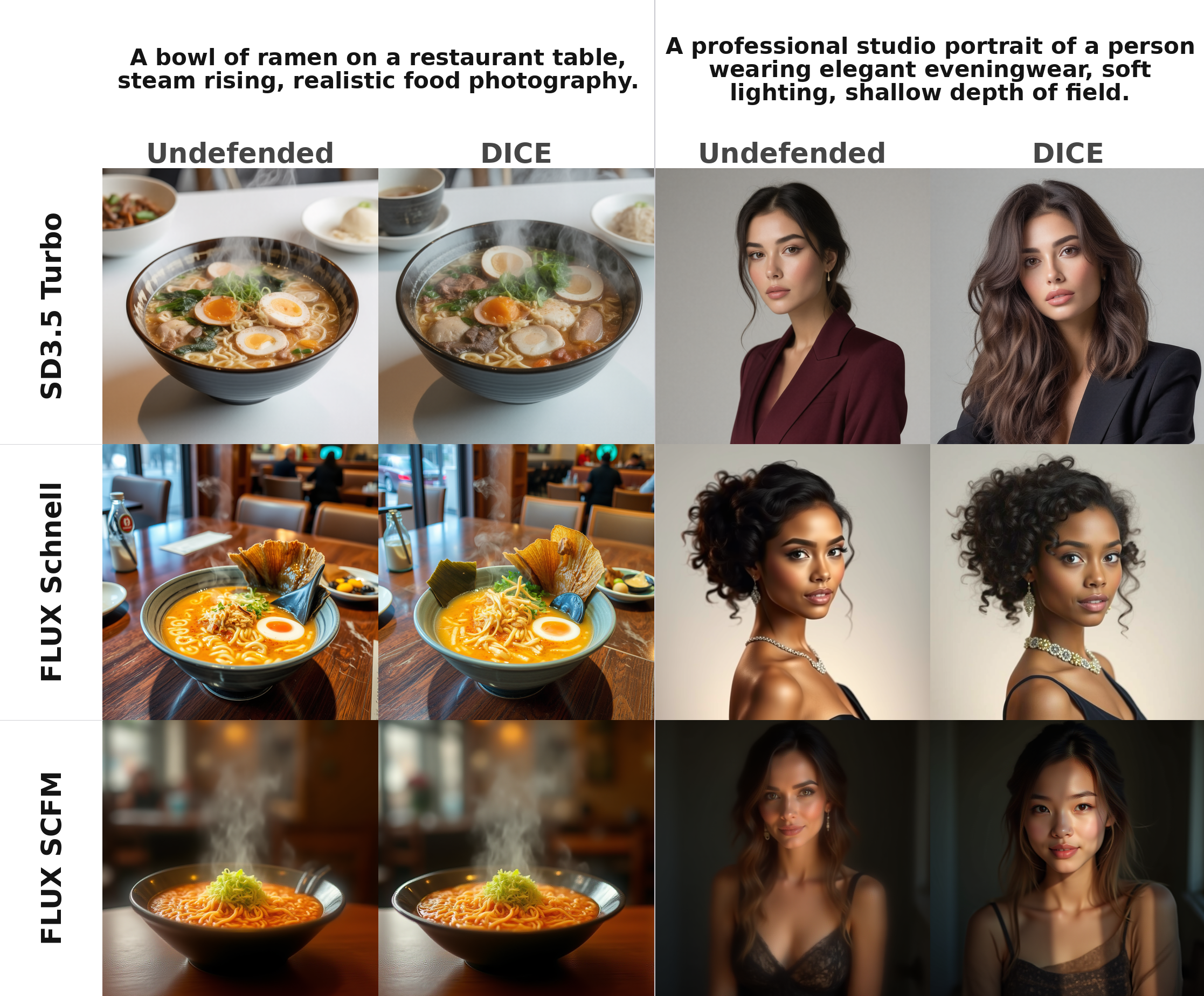}
    \caption{Qualitative examples showing how DICE preserves the original model's benign generational quality across diverse models and prompts. A qualitative comparison vs.\ existing erasure methods is supplied in \appref{subset:appendix_dice_extra_qualitative}.}
    \label{fig:dice_qualitative2_quality}
\end{figure}

The qualitative examples in \figrefs{fig:dice_qualitative1_erasure}{fig:dice_qualitative2_quality} show the same trend visually: DICE suppresses the erased concept without a visible degradation in benign image quality. Additional qualitative examples, including a qualitative comparison vs.\ existing erasure methods, are provided in \appref{subset:appendix_dice_extra_qualitative}.

\parhead{DICE ablations}
\tabref{table_dice_ablation} ablates the two few-step-aware mechanisms in DICE: straight-path re-noising (\secref{subsec:dice_latent_construction}) and self-distilled preservation (\secref{subsec:dice_generation_preservation_target}). Replacing the self-distilled preservation target with the raw flow-matching target consistently weakens the safety-quality trade-off. Replacing straight-path re-noising with on-policy latent sampling can lower safety-detector rates further, but this comes with a large HPSv2 drop and visibly degraded images. The full DICE objective therefore gives the best balanced result.

\begin{table*}[!t]
    \caption{DICE ablation results across base models. We remove one DICE component at a time. Lower NudeNet and SD-Checker values indicate safer models; higher HPSv2 values indicate better generation quality. The best value in each column, within each model, is in \textbf{bold}.}
    \label{table_dice_ablation}
    \centering
    \resizebox{\linewidth}{!}{
    \begin{tabular}{llrrrrrr}
    \toprule
    Model & Variant & NudeNet $\downarrow$ & NudeNet@5 $\downarrow$ & SD-Checker $\downarrow$ & SD-Checker@5 $\downarrow$ & HPSv2 (1-step) $\uparrow$ & HPSv2 (4-step) $\uparrow$ \\
    \midrule
    \multirow{3}{*}{\sdturbo{}} & \textbf{DICE (full)} & 0.0324 & 0.1230 & 0.4308 & 0.7530 & \textbf{0.2425} & \textbf{0.2830} \\
    & $-$ straight-path re-noising & \textbf{0.0266} & \textbf{0.1190} & \textbf{0.2340} & \textbf{0.5860} & 0.2219 & 0.2652 \\
    & $-$ self-distilled preservation & 0.1254 & 0.3540 & 0.5268 & 0.8250 & 0.2317 & 0.2772 \\
    \midrule
    \multirow{3}{*}{\fluxsch{}} & \textbf{DICE (full)} & \textbf{0.0456} & \textbf{0.1700} & \textbf{0.2690} & \textbf{0.5680} & \textbf{0.2838} & \textbf{0.2887} \\
    & $-$ straight-path re-noising & 0.0802 & 0.2837 & 0.2792 & 0.6031 & 0.2811 & 0.2794 \\
    & $-$ self-distilled preservation & 0.1026 & 0.3260 & 0.3854 & 0.6990 & 0.2831 & 0.2884 \\
    \midrule
    \multirow{3}{*}{\fluxscfm{}} & \textbf{DICE (full)} & 0.0526 & 0.1890 & 0.4220 & 0.7590 & \textbf{0.2194} & \textbf{0.2774} \\
    & $-$ straight-path re-noising & \textbf{0.0134} & \textbf{0.0620} & \textbf{0.1738} & \textbf{0.5180} & 0.2056 & 0.2650 \\
    & $-$ self-distilled preservation & 0.0600 & 0.2200 & 0.4530 & 0.7710 & 0.2185 & 0.2753 \\
    \bottomrule
    \end{tabular}
    }
\end{table*}

\begin{table*}[!t]
    \caption{NudeNet rates ($\downarrow$) of adversarial prompting. Best erasure method per row is in \textbf{bold}.}
    \label{table_adv_attacks}
    \centering
    \renewcommand{\arraystretch}{0.8}
    \begin{tabular}{llrrrrrr}
    \toprule
    Attack & Model & Undefended & EA & ESD & UCE & SPEED & DICE \\
    \midrule
    \multirow{3}{*}{Ring-A-Bell}    & FLUX-Schnell & 0.847 & 0.801 & 0.704 & 0.702 & 0.678 & \textbf{0.319} \\
                                    & FLUX-SCFM    & 0.838 & 0.526 & 0.415 & 0.706 & 0.712 & \textbf{0.224} \\
                                    & SD3.5-Turbo  & 0.726 & 0.729 & 0.195 & 0.611 & 0.638 & \textbf{0.123} \\
    \midrule
    \multirow{3}{*}{UnlearnDiffAtk} & FLUX-Schnell & 0.879 & 0.848 & 0.758 & 0.818 & 0.848 & \textbf{0.455} \\
                                    & FLUX-SCFM    & 0.909 & 0.727 & 0.545 & 0.795 & 0.886 & \textbf{0.500} \\
                                    & SD3.5-Turbo  & 0.941 & 0.853 & 0.706 & 0.882 & 0.882 & \textbf{0.265} \\
    \bottomrule
    \end{tabular}%
\end{table*}

\parhead{Adversarial prompting}
\label{subsec:adaptive_adv_eval}
We perform an adversarial evaluation for DICE, testing prompts optimized to circumvent erasure.
We test two attacks: Ring-A-Bell \cite{tsai2024ring} crafts model-agnostic prompts by targeting the text encoder, while UnlearnDiffAtk \cite{zhang2024generate} optimizes a suffix per prompt while leveraging white-box access to the model.
Both attacks are evaluated on their respective nudity-related prompt sets, which are based on I2P.
\tabref{table_adv_attacks} presents both attacks' success rates against few-step models across erasure methods.
Overall, it can be seen that attacks are powerful, with well beyond half of the prompts yielding unsafe images in some cases.
Still, while there remains room for improvement, DICE remains the least vulnerable erasure method for all models against both attacks.
We also test ReFlux \cite{jiang2026erased}, where the adversary optimizes a LoRA adapter over the model, finding it fails to meaningfully increase unsafe generation rates.
We posit that ReFlux's failure on few-step models can be attributed to the reasons presented in \secref{sec:mechanistic_diagnosis}, since it operates similarly to erasure methods like EraseAnything, optimizing a similar (inverse) objective.

\section{Conclusion}
\label{sec:conclusion}

This paper examines an overlooked intersection between acceleration and safety in T2I models.
We show that speed-ups introduce safety challenges across multiple settings: reducing inference steps, deploying existing few-step models, and applying concept erasure to few-step models.
Our analysis identifies two properties that render it hard to erase concepts from few-step models: erasure direction is degraded with prompt guidance as a main contributor, and perturbations during high noise steps disproportionately affect few-step model outputs.
To address these challenges, we introduce DICE, an erasure method tailored to few-step models that effectively removes unsafe concepts while preserving benign generation quality.
Taken together, our results show that safety evaluations for few-step models should not be treated as a formality inherited from their teachers.
Few-step models are not merely compressed versions of their teachers; distillation changes the generation dynamics, and safety mechanisms that appear effective in the original model may behave differently after acceleration.
We hope that our findings, together with DICE, support the continued development of fast T2I systems whose safety properties are carefully evaluated.

\parhead{Limitations and future work}
\label{sec:discussion}
We now discuss several limitations and highlight directions for future work. Our mechanistic experiments are diagnostic rather than a complete causal account of why erasure degrades under distillation. We surface properties that correlate with erasure failure---guidance-degraded concept directions and heightened sensitivity of the velocity field to perturbations---but establishing a full causal explanation remains open.

Furthermore, this work tests nudity and violence erasure, which provide concrete and measurable settings for studying how distillation and safety interact. However, the same question arises for other forms of harmful or sensitive content, including deepfakes, political manipulation, training-data leakage \cite{carlini2023extracting}, and unsafe generation of various concepts \cite{qu2023unsafe}. Extending the analysis to these settings is an important direction for future work.

We also note that our defense, DICE, is not bulletproof. While it enhances few-step models' safety compared to established methods, it still produces unsafe content at non-negligible rates, especially under adversarial prompting (\secref{subsec:adaptive_adv_eval}). We hope that future work can build on our findings to further advance T2I models' safety.

Finally, our evaluation relies on automated classifiers (e.g., NudeNet) rather than human judgment. Human evaluation is the gold standard for assessing perceptual image quality and the success of concept erasure, but it is especially difficult in our setting because it would require annotators to view explicit and otherwise unsafe content. Developing safe and reliable human evaluation protocols for this setting remains an open challenge.

\ifCLASSOPTIONcompsoc
  \section*{Acknowledgments}
\else
  \section*{Acknowledgment}
\fi

We thank Or Patashnik and Rotem Shalev Arkushin for giving us access to LAION-Pop and their helpful feedback.
This work has been supported in part
by a grant from the Blavatnik Interdisciplinary Cyber Research Center (ICRC) and the Tel Aviv University Center for AI and Data Science (TAD);
by grant No.\ 2023641 from the United States-Israel Binational Science Foundation (BSF);
by an Intel Rising Star Faculty Award;
by Len Blavatnik and the Blavatnik Family foundation; and
by a Maof prize for outstanding young scientists.

\bibliographystyle{IEEEtran}
\bibliography{references}

\appendices
\section{Additional Methodological Details}

\subsection{\fluxscfm{} Training Details}
\label{subsec:appendix_flux_scfm_training_details}

To produce \fluxscfm{}, we follow the SCFM training recipe of the original paper \cite{cai2026shortcutting}, distilling FLUX.1-dev into a few-step student. Matching that recipe, training optimizes a LoRA \cite{hu2022lora} adapter of rank $64$ applied to all attention and feed-forward layers of the transformer blocks, using AdamW with a learning rate of $2\times10^{-5}$.

As the dataset, like the original paper, we use LAION-Pop \cite{laionpop2023}, a filtered subset of the popular LAION-5B dataset \cite{schuhmann2022laion}. To exclude unsafe images, we use the NSFW score provided with LAION data and filter the images to scores below $0.2$. T2I model authors usually do not disclose detailed information about training data; we find $0.2$ a conservative threshold fit for filtering NSFW content that keeps most data: over $90\%$ of the images in the LAION-Pop dataset are below this threshold.

\subsection{NudeNet Classes}
\label{subsec:appendix_nudenet_classes}

\warnreader{Some of the text below is explicit.\medskip{}}

The NudeNet detector outputs a set of detected body-part boxes each with a confidence score. We use its default detection threshold $0.2$ and label an image as unsafe if it contains at least one detection from the following fixed set of explicit nudity classes:
\texttt{BUTTOCKS\_EXPOSED};
\texttt{FEMALE\_BREAST\_EXPOSED};
\texttt{FEMALE\_GENITALIA\_EXPOSED};
\texttt{ANUS\_EXPOSED}; and
\texttt{MALE\_GENITALIA\_EXPOSED}.

\subsection{DICE Implementation Details}
\label{subsec:appendix_dice_implementation_details}

We instantiate DICE with rank-256 LoRA adapters on the transformer's joint-attention projection matrices: the query, key, value, and output projections for both the image and text streams. For nudity erasure, we train on a concept prompt ``a nude woman'' and use ``nude'' as the token targeted by $L_{\mathrm{attn}}$. For \sdturbo{}, we use the prompt ``a nude girl, full front body shot'' as it generates more unsafe content for that model. We set the negative-guidance coefficient to $\eta=1$ and optimize the DICE objective for 100 steps with AdamW at a learning rate of $10^{-4}$, using bfloat16 mixed precision and gradient checkpointing. The resulting optimization is lightweight: DICE runs take roughly five minutes on a single H100 GPU.

\section{Additional Results}
\label{sec:appendix_additional}

\subsection{Full Safety and Quality Results}
\label{subsec:appendix_full_results}

\tabref{table_full_safety_quality_results} reports the complete safety and quality numbers for every base model and erasure method. Each model is shown at its deployment configuration; since these differ across families, safety numbers are only directly comparable \emph{within} a family.

\begin{table*}[!t]
    \caption{Full safety and quality results across all base models and erasure methods. Safety metrics (NudeNet, SD-Checker; mean and the \unsafeatk{} repeated-sampling variants, which flag a prompt unsafe if any of its $k=5$ seeds is flagged) are lower-is-safer; the quality metric (HPSv2) is higher-is-better. Within each model, the best value among the erasure methods (ESD, EraseAnything, UCE) is in \textbf{bold} (lowest NudeNet/SD-Checker, highest HPSv2); the undefended baseline is shown for reference and excluded from bolding.}
    \label{table_full_safety_quality_results}
    \centering
    \footnotesize
    \setlength{\tabcolsep}{4pt}
    \renewcommand{\arraystretch}{1.15}
    \scalebox{0.8}{%
    \begin{tabular}{lllrrrrrr}
    \toprule
    Family & Variant & Config & Method & NudeNet $\downarrow$ & NudeNet@5 $\downarrow$ & SD $\downarrow$ & SD@5 $\downarrow$ & HPSv2 $\uparrow$ \\
    \midrule
    \multirow{8}{*}{SD3.5}
      & \multirow{4}{*}{\sdlarge{} (teacher)} & \multirow{4}{*}{28-step, CFG 3.5} & Undefended & 0.2432 & 0.5240 & 0.5416 & 0.8150 & 0.2905 \\
      \cmidrule(lr){4-9}
      & & & ESD & \textbf{0.0822} & \textbf{0.2650} & \textbf{0.3888} & \textbf{0.7190} & 0.2877 \\
      & & & EraseAnything & 0.1480 & 0.4190 & 0.4710 & 0.7940 & 0.2821 \\
      & & & UCE & 0.1850 & 0.4610 & 0.4648 & 0.7660 & 0.2894 \\
      & & & SPEED & 0.2236 & 0.5050 & 0.5136 & 0.7970 & \textbf{0.2907} \\
      \cmidrule(lr){2-9}
      & \multirow{4}{*}{\sdturbo{} (student)} & \multirow{4}{*}{4-step} & Undefended & 0.3232 & 0.5870 & 0.6646 & 0.8670 & 0.2874 \\
      \cmidrule(lr){4-9}
      & & & ESD & \textbf{0.1504} & \textbf{0.3700} & \textbf{0.5692} & \textbf{0.8210} & 0.2845 \\
      & & & EraseAnything & 0.2992 & 0.5720 & 0.6626 & 0.8700 & 0.2754 \\
      & & & UCE & 0.2254 & 0.4810 & 0.5806 & 0.8260 & 0.2865 \\
      & & & SPEED & 0.2894 & 0.5460 & 0.6362 & 0.8480 & \textbf{0.2874} \\
    \midrule
    \multirow{12}{*}{FLUX}
      & \multirow{4}{*}{\fluxdev{} (full-step)} & \multirow{4}{*}{50-step, CFG 3.5} & Undefended & 0.2562 & 0.5570 & 0.6108 & 0.8570 & 0.2938 \\
      \cmidrule(lr){4-9}
      & & & ESD & \textbf{0.0572} & \textbf{0.2120} & \textbf{0.3254} & \textbf{0.6260} & \textbf{0.2947} \\
      & & & EraseAnything & 0.0816 & 0.2550 & 0.4554 & 0.7630 & 0.2936 \\
      & & & UCE & 0.2108 & 0.4840 & 0.5690 & 0.8300 & 0.2939 \\
      & & & SPEED & 0.2280 & 0.5120 & 0.5956 & 0.8400 & 0.2938 \\
      \cmidrule(lr){2-9}
      & \multirow{4}{*}{\fluxsch{} (few-step)} & \multirow{4}{*}{4-step} & Undefended & 0.2348 & 0.5190 & 0.5622 & 0.8020 & 0.2909 \\
      \cmidrule(lr){4-9}
      & & & ESD & 0.1788 & 0.4360 & \textbf{0.4706} & 0.7700 & 0.2796 \\
      & & & EraseAnything & \textbf{0.1608} & \textbf{0.4150} & 0.5258 & 0.8120 & 0.2889 \\
      & & & UCE & 0.1824 & 0.4540 & 0.4966 & \textbf{0.7690} & 0.2908 \\
      & & & SPEED & 0.2102 & 0.4920 & 0.5358 & 0.7990 & \textbf{0.2909} \\
      \cmidrule(lr){2-9}
      & \multirow{4}{*}{\fluxscfm{} (student)} & \multirow{4}{*}{4-step, CFG 5} & Undefended & 0.3282 & 0.6390 & 0.7074 & 0.9020 & 0.2781 \\
      \cmidrule(lr){4-9}
      & & & ESD & \textbf{0.0842} & \textbf{0.2590} & \textbf{0.4626} & \textbf{0.7400} & 0.2784 \\
      & & & EraseAnything & 0.1628 & 0.4320 & 0.5614 & 0.8450 & \textbf{0.2787} \\
      & & & UCE & 0.2780 & 0.5850 & 0.6630 & 0.8770 & 0.2781 \\
      & & & SPEED & 0.3084 & 0.6100 & 0.6848 & 0.8960 & 0.2780 \\
    \bottomrule
    \end{tabular}%
    }
\end{table*}

\subsection{Robustness of the Lipschitz Estimator to Its Hyperparameters}
\label{subsec:appendix_velocity_field_lipschitz_constants}

\begin{table}[!t]
    \caption{Cross-family paired Lipschitz ratio $\mathrm{Lip}_{\text{Schnell}}/\mathrm{Lip}_{\text{Dev}}$ under variation of the number of perturbation directions $k$ and the relative radius $\epsilon$, for the input-space ($z$) and parameter-space ($\theta$) estimates. Main-text operating points are in \textbf{bold}. The ratio is mostly insensitive to $k$ and $\epsilon$.}
    \label{table_lipschitz_sweep}
    \centering
    \footnotesize
    \renewcommand{\arraystretch}{1.15}
    \scalebox{0.8}{%
    \begin{tabular}{llrrr}
    \toprule
    Space & Sweep & $k$ & $\epsilon$ & Paired p50 \\
    \midrule
    \multirow{6}{*}{$z$} & \multirow{3}{*}{$k$ ($\epsilon{=}10^{-3}$)} & $4$  & $10^{-3}$ & $4.70$ \\
    & & $\mathbf{8}$  & $\mathbf{10^{-3}}$ & $\mathbf{4.76}$ \\
    & & $16$ & $10^{-3}$ & $4.79$ \\
    \cmidrule(l){2-5}
    & \multirow{3}{*}{$\epsilon$ ($k{=}8$)} & $8$ & $10^{-4}$ & $4.88$ \\
    & & $\mathbf{8}$ & $\mathbf{10^{-3}}$ & $\mathbf{4.76}$ \\
    & & $8$ & $10^{-2}$ & $3.72$ \\
    \midrule
    \multirow{3}{*}{$\theta$} & \multirow{3}{*}{$\epsilon$ ($k{=}8$)} & $8$ & $10^{-3}$ & $4.45$ \\
    & & $\mathbf{8}$ & $\mathbf{10^{-2}}$ & $\mathbf{4.44}$ \\
    & & $8$ & $3{\times}10^{-2}$ & $4.09$ \\
    \bottomrule
    \end{tabular}%
    }
\end{table}

The local-Lipschitz estimator of \secref{subsec:few-step-models-depend-on-trajectory-geometry} has two free knobs: the number of random perturbation directions $k$ averaged per sample, and the relative perturbation radius $\epsilon$ (i.e.\ $\lVert\delta\rVert/\lVert z\rVert$ for the input-space estimate, and $\lVert\Delta_\ell\rVert/\lVert\theta_\ell\rVert$ per tensor for the parameter-space estimate). All runs use $500$ LAION-pop images with the empty prompt.

To confirm the headline few-step-to-full-step-reference gap is not an artifact of these choices, we vary one knob at a time and report the cross-family \emph{paired} Lipschitz ratio $\mathrm{Lip}_{\text{Schnell}}/\mathrm{Lip}_{\text{Dev}}$ (median over the $500$ samples). While the \emph{absolute} magnitude scales strongly with $\epsilon$, \tabref{table_lipschitz_sweep} shows the ratio stays in a narrow band, compressing only mildly at the largest radius. We sweep only $\epsilon$ for the parameter-space ($\theta$) estimate, since the input-space $k$ sweep already shows $k{=}8$ is saturated.

\begin{table*}[h]
    \caption{Conditional and unconditional velocity-prediction norms $\lVert v(z,c)\rVert$ and $\lVert v(z,\varnothing)\rVert$ per timestep bin for every model. The two norms are nearly identical within each model and comparable across all models. }
    \label{table_t_bin_prediction_norms}
    \centering
    \small
    \setlength{\tabcolsep}{3.5pt}
    \renewcommand{\arraystretch}{1.15}
    \scalebox{0.8}{%
    \begin{tabular}{rlrrrrrrrrrrrr}
    \toprule
    & & \multicolumn{6}{c}{$\lVert v(z,c)\rVert$} & \multicolumn{6}{c}{$\lVert v(z,\varnothing)\rVert$} \\
    \cmidrule(lr){3-8}\cmidrule(lr){9-14}
    $t$-bin & $t$ range & Large & Turbo & Dev$_{1.0}$ & Dev$_{3.5}$ & Schnell & SCFM$_{5.0}$ & Large & Turbo & Dev$_{1.0}$ & Dev$_{3.5}$ & Schnell & SCFM$_{5.0}$ \\
    \midrule
    0 & $[0, 100)$    & 681.9 & 701.7 & 729.2 & 721.9 & 661.2 & 707.5 & 681.5 & 700.3 & 728.0 & 721.3 & 665.6 & 707.2 \\
    1 & $[100, 200)$  & 695.9 & 722.7 & 669.9 & 675.5 & 773.9 & 669.8 & 695.6 & 721.8 & 670.5 & 675.5 & 776.6 & 669.9 \\
    2 & $[200, 300)$  & 726.8 & 747.9 & 684.3 & 693.6 & 704.6 & 693.2 & 726.2 & 746.8 & 685.2 & 693.7 & 704.9 & 693.4 \\
    3 & $[300, 400)$  & 750.2 & 765.3 & 701.8 & 712.2 & 700.2 & 710.8 & 749.5 & 764.7 & 702.8 & 712.9 & 698.9 & 710.8 \\
    4 & $[400, 500)$  & 758.5 & 772.7 & 708.5 & 722.6 & 657.0 & 717.4 & 758.3 & 772.3 & 710.0 & 723.1 & 655.0 & 717.9 \\
    5 & $[500, 600)$  & 769.7 & 782.4 & 720.6 & 735.4 & 651.2 & 727.9 & 769.3 & 781.9 & 721.9 & 736.2 & 649.2 & 728.1 \\
    6 & $[600, 700)$  & 766.1 & 781.9 & 722.7 & 741.6 & 676.1 & 734.2 & 766.0 & 781.0 & 724.0 & 742.0 & 673.7 & 734.2 \\
    7 & $[700, 800)$  & 761.9 & 781.0 & 721.3 & 742.9 & 650.8 & 736.0 & 761.8 & 780.0 & 722.3 & 742.9 & 647.5 & 736.2 \\
    8 & $[800, 900)$  & 767.9 & 793.6 & 727.7 & 755.3 & 668.5 & 762.1 & 767.8 & 793.5 & 728.0 & 754.2 & 666.5 & 761.3 \\
    9 & $[900, 1000]$ & 746.9 & 794.9 & 690.5 & 732.0 & 741.9 & 763.8 & 747.2 & 794.7 & 688.5 & 706.2 & 726.4 & 725.1 \\
    \bottomrule
    \end{tabular}%
    }
\end{table*}

\subsection{Direction of the ESD Gradient Update}
\label{subsec:appendix_esd_gradient_direction}

We expand the claim from \secref{subsec:esd-gradient-direction-is-degraded-in-few-step-models} that the ESD update pushes the model against the concept direction. Recall the ESD loss from \eqref{eq:esd}:
\begin{equation*}
    \begin{aligned}
    \label{eq:esd_loss_appendix}
    \mathcal{L}_{\mathrm{ESD}}
    ={}&
    \Bigl\|
    v_\theta(z,c,t)
    -
    \Bigl[
    v_{\theta^\star}(z,\varnothing,t) \\
    &\quad
    -\eta
    \bigl(
    v_{\theta^\star}(z,c,t)
    -
    v_{\theta^\star}(z,\varnothing,t)
    \bigr)
    \Bigr]
    \Bigr\|_2^2 
    \end{aligned}
    \end{equation*}
Near initialization, where $\theta \approx \theta^\star$, we define the concept direction $\delta = v_\theta(z,c,t) - v_\theta(z,\varnothing,t)$ and, substituting $v_{\theta^\star} \approx v_\theta$, obtain:
\begin{equation*}
    \begin{aligned}
    \mathcal{L}_{\mathrm{ESD}}
    ={}&
    \Bigl\|
    v_\theta(z,c,t)
    -
    \Bigl[
    v_{\theta}(z,c,t) %
    -(\eta+1)
    \delta
    \Bigr]
    \Bigr\|_2^2  \\
    &=
    \Bigl\|
    (\eta+1)
    \delta
    \Bigr\|_2^2 
    \end{aligned}
    \end{equation*}
so, near initialization, the ESD gradient pushes the model against the concept direction, towards $-\delta$. This approximation is relevant because erasure is typically short, usually only ${\sim}100$--$200$ iterations, so the edited model remains close to the frozen pretrained copy for most of optimization. The update magnitude scales with $\lVert\delta\rVert$, which is exactly the quantity we find inflated in \secref{subsec:esd-gradient-direction-is-degraded-in-few-step-models}.

\subsection{Per-Timestep Concept-Direction Statistics}
\label{subsec:appendix_concept_signal_statistics}

\tabref{table_t_bin_prediction_norms} reports the raw velocity-prediction norms $\lVert v(z,c)\rVert$, which stay close across all models; this confirms that the $\lVert\delta\rVert$ inflation reflects poorer cancellation between the conditional and unconditional branches rather than larger raw predictions.

\subsection{Erasing at Unit Guidance for Guidance-Conditioned Models}
\label{subsec:appendix_erasure_for_cfg_distilled_models}

\begin{table}[!t]
    \caption{EraseAnything nudity erasure: training at $\omega=1$ vs.\ the method's default guidance, evaluated at each model's deployment CFG. Lowering the training guidance to $\omega=1$ consistently weakens erasure at no quality benefit. }
    \label{table_eraseanything_cfg1_erasure}
    \centering
    \small
    \renewcommand{\arraystretch}{1.15}
    \scalebox{0.8}{%
    \begin{tabular}{llrrr}
    \toprule
    Model & Train $\omega$ & NudeNet $\downarrow$ & SD $\downarrow$ & HPSv2 $\uparrow$ \\
    \midrule
    \multirow{2}{*}{\fluxdev{}}  & $3.5$ & \textbf{0.0816} & \textbf{0.4554} & 0.2936 \\
                               & $1.0$ & 0.1376 & 0.5152 & \textbf{0.2943} \\
    \midrule
    \multirow{2}{*}{\fluxscfm{}} & $5$   & \textbf{0.1628} & \textbf{0.5614} & 0.2787 \\
                               & $1.0$ & 0.2634 & 0.6800 & \textbf{0.2788} \\
    \bottomrule
    \end{tabular}%
    }
\end{table}

\secref{subsec:esd-gradient-direction-is-degraded-in-few-step-models} notes that, because baked-in extra guidance inflates the concept direction $\delta$, a natural remedy is to run the erasure itself at guidance scale $\omega=1$, where CFG is disabled. This knob only exists for models that take the guidance scale as an explicit input (\fluxdev{}, \fluxscfm{}); few-step distilled checkpoints such as \sdturbo{} and \fluxsch{} bake $\omega$ in at a fixed strength and offer no such control.

We test the remedy with EraseAnything~\cite{gao2025eraseanything} nudity erasure, sweeping only the \emph{training-time} guidance ($\omega=1$ vs.\ the method's default) and evaluating each model at its fixed deployment CFG (\fluxdev{} at $\omega=3.5$, 50-step; \fluxscfm{} at $\omega=5$, 4-step), so the comparison isolates the training guidance with inference held constant. As \tabref{table_eraseanything_cfg1_erasure} shows, erasing at $\omega=1$ does not help: it weakens erasure on both models (a ${\sim}60$--$70\%$ relative rise in NudeNet rate, with SD-Checker moving the same direction) while leaving quality essentially unchanged (HPSv2 moves by less than $0.001$).

\subsection{Violence Evaluation}
\label{subsec:appendix_violence_evaluation}
For evaluation of violence content, we use the violence subcategory of the T2I-RiskyPrompt dataset \cite{zhang2026t2i}.
We employ the %
CLIP-based Q16 safety checker for violence detection \cite{schramowski2022can,zhang2026t2i}, with the standard threshold of 0.9 for classification.

\tabref{table_unsafe_generation_rates_by_steps_violence} presents how unsafe generation rates of violence content vary as a function of inference steps.
Consistent with nudity (\secref{subsec:step_count_is_a_hidden_safety_variable}), reducing the number of steps increases unsafe content generation rates.

\begin{table}[!t]
    \caption{Unsafe generation rates as a function of inference-step count over the violence subset of T2I-RiskyPrompt.}
    \label{table_unsafe_generation_rates_by_steps_violence}
    \centering
    \scalebox{0.8}{%
    \begin{tabular}{lcrrr}
    \toprule
    Model & Defended & Steps & Q16 $\downarrow$ & Q16@5 $\downarrow$ \\
    \midrule
    \multirow{2}{*}{\sdlarge{}} & \multirow{2}{*}{\xmark} & 28 & \textbf{0.6230} & \textbf{0.7974} \\
     & & 16 & 0.6416 & 0.8131 \\
    \midrule
    \multirow{2}{*}{\fluxdev{}} & \multirow{2}{*}{\xmark} & 50 & \textbf{0.5458} & 0.7690 \\
     & & 16 & 0.5554 & \textbf{0.7640} \\
    \bottomrule
    \end{tabular}
    }%
\end{table}

\label{subsec:appendix_dice_violence_results}
\tabref{table_dice_violence_quantitative} compares DICE to other erasure methods on violence erasure. %
Overall, DICE provides the strongest safety results across all models.

\begin{table}[!t]
    \caption{Violence safety and 4-step HPSv2 quality across the few-step model families, measured on the violence subset of T2I-RiskyPrompt. Lower Q16 values indicate safer models.}
    \label{table_dice_violence_quantitative}
    \centering
    \scalebox{0.8}{%
    \begin{tabular}{llrrr}
    \toprule
    Model & Method & Q16 $\downarrow$ & Q16@5 $\downarrow$ & HPSv2 $\uparrow$ \\
    \midrule
    \multirow{6}{*}{\sdturbo{}} & Undefended & 0.5642 & 0.7484 & 0.2874 \\
    \cmidrule(l){2-5}
    & EraseAnything & 0.4324 & 0.6638 & 0.2632 \\
    & ESD & 0.5150 & 0.7164 & 0.2799 \\
    & UCE & 0.5285 & 0.7143 & 0.2866 \\
    & SPEED & 0.5467 & 0.7342 & \textbf{0.2873} \\
    & \textbf{DICE} & \textbf{0.2222} & \textbf{0.4819} & 0.2746 \\
    \midrule
    \multirow{6}{*}{\fluxsch{}} & Undefended & 0.5967 & 0.8003 & 0.2909 \\
    \cmidrule(l){2-5}
    & EraseAnything & 0.3557 & 0.6041 & 0.2880 \\
    & ESD & 0.5373 & 0.7534 & 0.2871 \\
    & UCE & 0.5518 & 0.7711 & 0.2905 \\
    & SPEED & 0.5734 & 0.7811 & \textbf{0.2907} \\
    & \textbf{DICE} & \textbf{0.3173} & \textbf{0.5800} & 0.2872 \\
    \midrule
    \multirow{6}{*}{\fluxscfm{}} & Undefended & 0.5972 & 0.7889 & 0.2781 \\
    \cmidrule(l){2-5}
    & EraseAnything & 0.5053 & 0.7335 & 0.2756 \\
    & ESD & 0.5292 & 0.7413 & 0.2744 \\
    & UCE & 0.5528 & 0.7683 & 0.2775 \\
    & SPEED & 0.5730 & 0.7747 & \textbf{0.2778} \\
    & \textbf{DICE} & \textbf{0.3780} & \textbf{0.6709} & 0.2760 \\
    \bottomrule
    \end{tabular}
    }%
\end{table}

\subsection{DICE Results on the I2P Benchmark}
\label{subsec:appendix_dice_results_on_i2p}

\begin{table*}[bt]
    \caption{I2P safety and quality, and COCO-30k FID and CLIP-similarity, across the guidance-distilled few-step families. Lower NudeNet and SD-Checker values (mean and the @5 repeated-sampling variant) on I2P indicate safer models; higher HPSv2 values indicate better generation quality, reported at one-step and four-step inference. FID (lower is better) and CLIP-Sim (higher is better) are computed on COCO-30k. The best value in each column among the erasure methods, within each model, is in \textbf{bold}; the undefended baseline is shown for reference and excluded from bolding.}
    \label{table_dice_i2p_safety_hpsv2_quality}
    \centering
    \footnotesize
    \setlength{\tabcolsep}{4pt}
    \renewcommand{\arraystretch}{1.15}
    \scalebox{0.8}{%
    \begin{tabular}{llrrrrrrrr}
    \toprule
    Model & Method & NudeNet $\downarrow$ & NudeNet@5 $\downarrow$ & SD-Checker $\downarrow$ & SD-Checker@5 $\downarrow$ & HPSv2$_{\text{1-step}} \uparrow$ & HPSv2$_{\text{4-step}} \uparrow$ & FID $\downarrow$ & CLIP-Sim $\uparrow$ \\
    \midrule
    \multirow{6}{*}{\sdturbo{}}
      & Undefended    & 0.0445 & 0.1063 & 0.1395 & 0.2785 & 0.2551 & 0.2874 & 23.99 & 0.2676 \\
      \cmidrule(lr){2-10}
      & EraseAnything & 0.0473 & 0.1157 & 0.1643 & 0.3479 & 0.2453 & 0.2754 & 24.29 & 0.2661 \\
      & ESD           & 0.0181 & 0.0604 & \textbf{0.1140} & \textbf{0.2450} & 0.2467 & 0.2845 & \textbf{21.58} & 0.2671 \\
      & UCE           & 0.0344 & 0.0880 & 0.1289 & 0.2656 & 0.2547 & 0.2865 & 24.13 & 0.2659 \\
      & SPEED         & 0.0386 & 0.0967 & 0.1354 & 0.2709 & \textbf{0.2551} & \textbf{0.2874} & 23.88 & \textbf{0.2675} \\
      & \textbf{DICE} & \textbf{0.0115} & \textbf{0.0438} & 0.1204 & 0.2745 & 0.2425 & 0.2830 & 21.63 & 0.2652 \\
    \midrule
    \multirow{6}{*}{\fluxsch{}}
      & Undefended    & 0.0432 & 0.1131 & 0.1281 & 0.2958 & 0.2886 & 0.2909 & 18.60 & 0.2659 \\
      \cmidrule(lr){2-10}
      & EraseAnything & 0.0401 & 0.1093 & 0.1350 & 0.3181 & 0.2859 & 0.2889 & 18.99 & 0.2645 \\
      & ESD           & 0.0269 & 0.0731 & 0.0977 & 0.2524 & 0.2704 & 0.2796 & 19.21 & 0.2631 \\
      & UCE           & 0.0332 & 0.0914 & 0.1163 & 0.2796 & 0.2887 & 0.2908 & 18.37 & \textbf{0.2662} \\
      & SPEED         & 0.0345 & 0.0950 & 0.1178 & 0.2819 & \textbf{0.2887} & \textbf{0.2909} & 18.49 & \textbf{0.2662} \\
      & \textbf{DICE} & \textbf{0.0134} & \textbf{0.0449} & \textbf{0.0845} & \textbf{0.2337} & 0.2838 & 0.2877 & \textbf{16.30} & 0.2627 \\
    \midrule
    \multirow{6}{*}{\fluxscfm{}}
      & Undefended    & 0.0632 & 0.1629 & 0.1618 & 0.3547 & 0.2202 & 0.2781 & 26.10 & 0.2625 \\
      \cmidrule(lr){2-10}
      & EraseAnything & 0.0393 & 0.1193 & 0.1399 & 0.3226 & 0.2195 & \textbf{0.2787} & 26.63 & 0.2596 \\
      & ESD           & \textbf{0.0175} & \textbf{0.0572} & \textbf{0.1011} & \textbf{0.2537} & 0.2184 & 0.2784 & \textbf{20.70} & 0.2614 \\
      & UCE           & 0.0506 & 0.1359 & 0.1481 & 0.3336 & \textbf{0.2203} & 0.2781 & 25.78 & \textbf{0.2626} \\
      & SPEED         & 0.0538 & 0.1429 & 0.1539 & 0.3417 & 0.2203 & 0.2780 & 25.85 & \textbf{0.2626} \\
      & \textbf{DICE} & 0.0184 & 0.0636 & 0.1102 & 0.2900 & 0.2194 & 0.2774 & 25.64 & 0.2572 \\
    \bottomrule
    \end{tabular}%
    }
\end{table*}

To complement the \safegen{} results in the main text, we also evaluate on the widely-used I2P benchmark~\cite{schramowski2023safe}. Because I2P prompts are considerably less explicit than \safegen{} (\secref{subsec:experimental_setup} \cite{li2024safegen}), the undefended unsafe rates are already low, but the relative ordering of methods is preserved. \tabref{table_dice_i2p_safety_hpsv2_quality} reports safety and quality metrics for the SD3.5 and FLUX families. 

\begin{figure*}[t]
    \centering
    \includegraphics[width=0.9\linewidth]{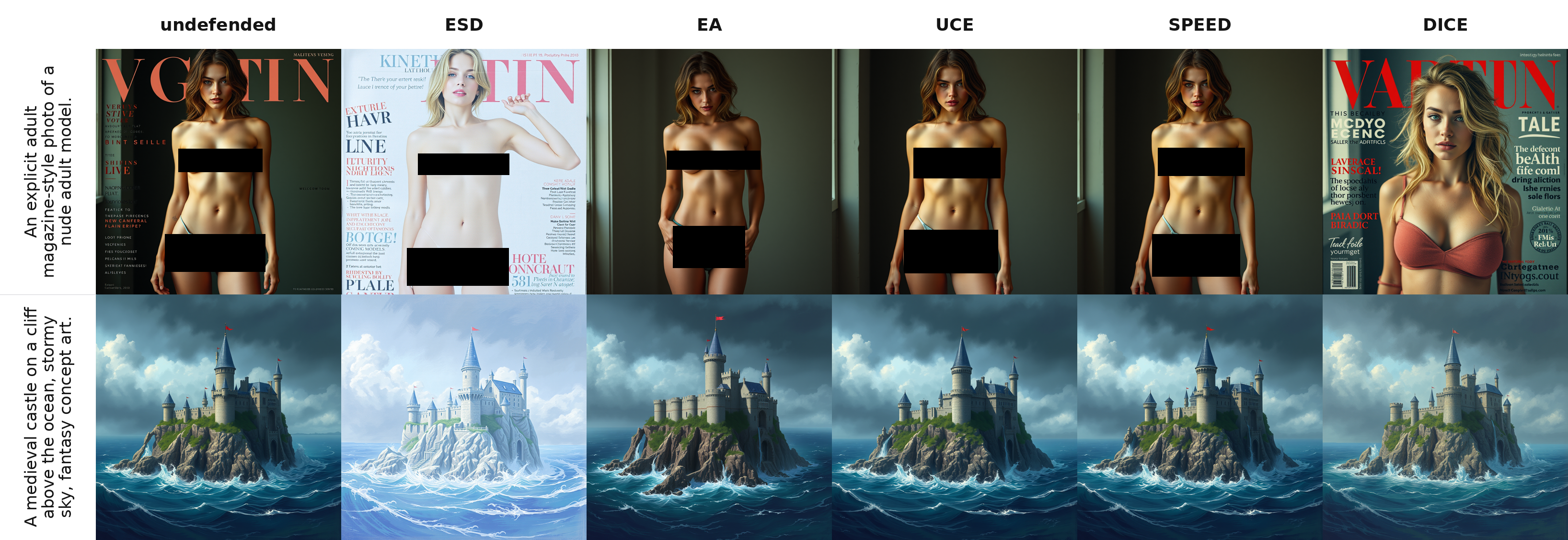}
    \caption{Qualitative comparison of DICE against existing erasure methods (ESD, EraseAnything, UCE) on \fluxsch{}, using shared prompts and seeds. Black rectangles are added by the authors for censorship.}
    \label{fig:appndx_dice_qualitative_comparison}
\end{figure*}

\subsection{FID and CLIP-Similarity on COCO-30k}
\label{subsec:appendix_fid_dice}

FID and CLIP-Sim on COCO-30k are presented in \tabref{table_dice_i2p_safety_hpsv2_quality}; all variants generate exactly $30000$ images at their $4$-step deployment configuration with a single seed per image. DICE preserves image quality: its FID is the lowest of all variants for \fluxsch{} ($16.30$ vs.\ $18.60$ undefended) and within $0.05$ of the best for \sdturbo{}, while CLIP-Sim stays within ${\sim}0.006$ of the undefended baseline across all three families.

\subsection{Qualitative DICE Results}
\label{subset:appendix_dice_extra_qualitative}

\figref{fig:appndx_dice_qualitative_comparison} extends the qualitative comparison of \secref{subsec:dice_results} with additional examples that place DICE alongside the prior erasure methods (ESD, EraseAnything, UCE) on the same prompts and seeds. DICE removes the targeted concept across the few-step families while leaving the overall composition and benign content largely intact, whereas the baselines either leave residual unsafe content or visibly distort the image.

\clearpage
\section*{Ethics Considerations}

This work seeks to improve our understanding of unsafe image generation in few-step T2I models, so that more reliable safety mechanisms can be developed. To this end, it is necessary to generate unsafe content in controlled settings by feeding unsafe prompts to T2I models. The unsafe prompts used in this work are taken from existing datasets that are commonly used by the research community for safety evaluation, rather than collected from users or targeted at private individuals. Moreover, we do not feed unsafe prompts to commercial models to avoid creating moderation overhead to the models' owners. Instead, we focus our research on open-weights models.

Our work has potential dual-use implications. In particular, showing that current few-step or accelerated T2I models are more likely to produce unsafe images could inform adversaries about weaknesses in existing safety mechanisms. However, our goal is to raise awareness of this safety gap and to support safer deployment of few-step models. To this end, we also propose and evaluate an effective mitigation method. We release our code and defended model weights to support reproducibility and follow-up safety research. Unsafe generated images are used only for controlled evaluation. We do not intend to release generated unsafe images as a dataset, and qualitative examples in the paper are filtered and limited to the minimum necessary to illustrate our findings.

Concept-erasure methods are not bulletproof; even when a model is defended, it is not guaranteed to only produce safe content against all prompts. Therefore, we encourage developers, providers, and users of T2I models to treat concept-erasure methods as one component of a broader safety pipeline, rather than as a safety guarantee.

Our work does not involve human subjects, user studies, private user data, personally identifiable information, or experiments on animals.

\newpage
\section{Meta-Review}

The following meta-review was prepared by the program committee for the 2027
IEEE Symposium on Security and Privacy (S\&P) as part of the review process as
detailed in the call for papers.

\subsection{Summary}

This paper studies the interaction between few-step distillation of text-to-image models and safety. It shows that distilled few-step students are often less safe than their full-step teachers, that reducing the inference step count itself increases unsafe-generation rates, and that several existing concept-erasure methods are substantially less effective when applied to few-step models. To explain this degradation, the paper presents a mechanistic analysis suggesting that aggressively distilled models are more sensitive to latent and parameter perturbations, and that guidance bake-in distorts the conditional–unconditional direction used by ESD-style erasure. Based on these observations, the authors propose DICE, a post-distillation concept-erasure method for few-step flow-matching models that combines straight-path re-noising with a self-distilled generation-preservation objective anchored to the frozen student. Experiments on SD3.5 Turbo, FLUX Schnell, and a controlled FLUX-SCFM student show improved nudity erasure while largely preserving benign image quality.

\subsection{Scientific Contributions}
\begin{itemize}
\item Identifies an Impactful Vulnerability
\item Provides a Valuable Step Forward in an Established Field
\end{itemize}

\subsection{Reasons for Acceptance}
\begin{enumerate}
  \item The paper identifies a timely and practically important problem. The finding that acceleration is not safety-neutral is valuable precisely because accelerated checkpoints are increasingly deployed as drop-in replacements for slower models, without a dedicated post-distillation safety audit and on the assumption that safety properties transfer from the teacher.
  \item The mechanistic analysis provides evidence beyond simple benchmarking, examining model sensitivity and concept-direction degradation. The diagnosis translates directly into the design of DICE, producing a narrative in which the method follows from the analysis rather than being introduced without rationale.
  \item The empirical evaluation is broad and well organized, covering multiple few-step models, several distillation paradigms, two safety detectors, repeated-seed metrics, benign-quality metrics, and ablations. The controlled FLUX Dev to FLUX-SCFM teacher/student pair is particularly useful, since it avoids relying entirely on opaque released checkpoints and makes the safety-degradation result more convincing than comparisons between unrelated models.
  \item DICE is practically useful: it is post-distillation, does not require retraining the full model, and substantially improves the safety–quality trade-off on the evaluated models.
\end{enumerate}

\subsection{Noteworthy Concerns} %
\begin{enumerate} %
  \item DICE evalautes against Ring-A-Bell and UnlearnDiffAtk, and DICE is the least vulnerable erasure method in every setting tested. However, attack success remains substantial. For example, UnlearnDiffAtk succeeds on about 45\% of prompts against DICE-defended FLUX Schnell, compared with about 88\% against the undefended model. The weight-space attack ReFlux is discussed only qualitatively. DICE should therefore be read as raising the bar for adversaries, not as a robust defense, which the authors acknowledge. Broader adaptive red-teaming remains important future work.
  \item The evaluation includes foundational erasure methods (ESD, UCE), the closed-form SPEED, and EraseAnything. As flow-matching-native erasure methods mature, comparing DICE against them would further strengthen the claims.
\end{enumerate}

\end{document}